\documentclass[sigconf]{acmart}
\AtBeginDocument{%
  \providecommand\BibTeX{{%
    \normalfont B\kern-0.5em{\scshape i\kern-0.25em b}\kern-0.8em\TeX}}}

\setcopyright{none}
\renewcommand\footnotetextcopyrightpermission[1]{} 

\usepackage{epsfig}
\usepackage{graphicx}
\usepackage{amsmath}
\usepackage{url}
\usepackage{enumitem}
\usepackage{wrapfig}
\usepackage{booktabs}       
\usepackage{subcaption}     
\usepackage{algorithm}
\usepackage{algpseudocode}
\usepackage{multirow}
\usepackage{xcolor}
\usepackage{amsthm}
\usepackage{marginnote}

\newtheorem{theorem}{Theorem}
\newtheorem{lemma}{Lemma}

\definecolor{myblue2}{HTML}{4682B4}
\definecolor{myred}{HTML}{FF5733}

\newcommand{\norm}[1]{\left\lVert #1 \right\rVert}

\begin{document}

\title{\textsc{Demon}: Improved Neural Network Training \\ with Momentum Decay}


\author{John Chen}
\affiliation{%
  \institution{Rice University}
  \country{USA}}
\email{johnchen@rice.edu}

\author{Cameron Wolfe}
\affiliation{%
  \institution{Rice University}
  \country{USA}}
\email{wolfe.cameron@rice.edu}

\author{Zhao Li}
\affiliation{%
  \institution{UTHealth}
  \country{USA}}
\email{Zhao.Li@uth.tmc.edu}

\author{Anastasios Kyrillidis}
\affiliation{%
  \institution{Rice University}
  \country{USA}}
\email{anastasios@rice.edu}

\renewcommand{\shortauthors}{Chen et al.}

\begin{abstract}
   Momentum is a widely used technique for gradient-based optimizers in deep learning. In this paper, we propose a decaying momentum (\textsc{Demon}) rule. 
   We conduct the first large-scale empirical analysis of momentum decay methods for modern neural network optimization, in addition to the most popular learning rate decay schedules. Across 28 relevant combinations of models, epochs, datasets, and optimizers, \textsc{Demon} achieves the highest number of Top-1 and Top-3 finishes at 39\% and 85\% respectively, almost doubling the second-placed learning rate cosine schedule at 17\% and 60\%, respectively.
   \textsc{Demon} also outperforms other widely used schedulers including, but not limited to, the learning rate step schedule, linear schedule, OneCycle schedule, and exponential schedule.
    Compared with the widely used learning rate step schedule, \textsc{Demon} is observed to be less sensitive to parameter tuning, which is critical to training neural networks in practice.
    Results are demonstrated across a variety of settings and architectures, including image classification, generative models, and language models.
    \textsc{Demon} is easy to implement, requires no additional tuning, and incurs almost no extra computational overhead compared to the vanilla counterparts. Code is readily available.
\end{abstract}



\maketitle

\pagestyle{plain} 



\section{Introduction}

\noindent \textbf{Motivation.} Deep Neural Networks (DNNs) have advanced the state-of-the-art in computer vision \cite{krizhevsky2012imagenet, he2016deep,ren2015faster}, natural language processing \cite{mikolov2013distributed,bahdanau2014neural, gehring2017convolutional} and speech recognition \cite{sak2014long,sercu2016very}, but have come with huge computation costs. A state-of-the-art language model can cost several million USD to train \cite{brown2020language, sharir2020the}. 
For most practitioners, even moderate tasks can be prohibitive in time and cost when the hyperparameter tuning process is taken into account, where it is typical to retrain models many times to achieve optimal performance.

\begin{table}
\centering
\begin{small}
\caption{\small Performance of different schedules ranked according to the \% of Top-1 or Top-3 finishes, out of a total of 28 experiments. Top-1 refers to the best performance for a particular model-dataset-base optimizer-epoch setting. Top-3 refers to among the best 3 performances. E.g. \texttt{RN20-CIFAR10-SGDM-300epochs} would be considered one experiment. \texttt{NCSN-CIFAR10}, \texttt{RN56-TINYIMAGENET} and \texttt{BERT$_{BASE}$-GLUE} not included due to limited runs.}
\begin{tabular}{c|cc} \toprule
    Method & Top-1 Performance& Top-3 Performance \\ \midrule
    OneCycle Momentum & 0\% & 0\% \\
    Linear Momentum & 0\% & 0\% \\
    Exp Momentum decay & 0\% & 0\% \\
    Cosine Momentum & 0\% & 3.57\% \\
    LR Exp decay & 7.14\% & 10.71\% \\
    LR Decay on Plateau & 7.14\% & 21.43\% \\
    OneCycle & 7.14\% & 25.00\% \\
    LR Linear Schedule & 10.71\% & 39.29\% \\
    LR Step Schedule & 10.71\% & 53.57\% \\
    LR Cosine Schedule & 17.86\% & 60.71\% \\ \midrule
    \textsc{Demon} & \textbf{39.29\%} & \textbf{85.71\%} \\
    \bottomrule
\end{tabular}
\label{tablepctresults}
\end{small}
\end{table}

In an effort to ease the cost of training DNNs, adaptive gradient-based methods \cite{duchi2011adaptive,zeiler2012adadelta,hinton2012neural,kingma2014adam,ma2018quasi} were devised. 
Cases exist where their use leads to degraded performance \cite{wilson2017marginal,shah2018minimum}, but this can be a result of poor hyperparameter tuning \cite{tune_opt, emp_opt, revis_opt, shah2018minimum}.
Currently, SGD with momentum (SGDM) and Adam \cite{kingma2014adam} remain among the most popular methods for training DNNs. 
Many state-of-the-art benchmarks in Computer Vision are achieved using SGDM and a curated learning rate step schedule \cite{krizhevsky2012imagenet,he2016deep,xie2017aggregated,zagoruyko2016wide,huang2017densely,ren2015faster,howard2017mobilenets}.
Meanwhile, variants of Adam are popular for training state-of-the-art language models \cite{devlin2019bert, brown2020language}.

For optimizers to achieve good performance, their hyperparameters must be tuned properly.
For example, slight changes in learning rate, learning rate decay, momentum, and weight decay (amongst others) can drastically alter performance.
One key component to hyperparameter tuning is the selection of a good learning rate, and possibly momentum, decay schedule.
Performance can vary substantially depending on this schedule, and, as we demonstrate, \emph{no one schedule is optimal.}
However, there is significant opportunity to improve upon existing schedules by fostering consistent state-of-the-art performance and robustness to hyperparameters across domains.
\begin{figure*}[th]
\begin{minipage}{0.3\textwidth}
    \begin{algorithm}[H]
    \centering
    \caption{\textsc{Demon} in SGDM}\label{alg:demon_sgdm}
    \begin{algorithmic}
    \vspace{0.2cm}
        \State \textbf{Parameters}: $T$ number of iterations, step size $\eta$, initial mom. $\beta_{\text{init}}$.       
        \State $v_0 = \theta_0 = 0$ or random.
        \For{$t = 0, \dots, T$}
        \State $\beta_t = \beta_{\text{init}} \cdot \tfrac{ \left(1 - \tfrac{t}{T} \right) }{ \left(1 - \beta_{\text{init}} \right) + \beta_{\text{init}} \left(1 - \frac{t}{T}\right)}$
        \State $\theta_{t+1} = \theta_t - \eta g_t + \beta_t v_t$
        \State $v_{t+1} = \beta_t v_t - \eta g_t$ 
        \EndFor   
        \vspace{0.2cm}
    \end{algorithmic}
\end{algorithm}
\end{minipage}
\hspace{0.1cm}
\begin{minipage}{0.34\textwidth}
\begin{algorithm}[H]
    \centering
    \caption{\textsc{Demon} in  Adam}\label{alg:demon_adam}
    \begin{algorithmic}
        \State \textbf{Parameters}: $T$, $\eta$, initial mom. $\beta_{\text{init}}$, $\beta_2$, $\varepsilon = 10^{-8}$. $v_0 = \theta_0 = 0$ or random.      
        \For{$t = 0, \dots, T$}
        \State $\beta_t = \beta_{\text{init}} \cdot \tfrac{ \left(1 - \tfrac{t}{T} \right) }{ \left(1 - \beta_{\text{init}} \right) + \beta_{\text{init}} \left(1 - \frac{t}{T}\right)}$
        \State $\mathcal{E}^{g \circ g}_{t+1} = \beta_2 \cdot \mathcal{E}^{g \circ g}_t + (1- \beta_2) \cdot (g_t \circ g_t)$
        \State $m_{t, i} = g_{t, 1} + \beta_{t} m_{t - 1, i}$
        \State $\theta_{t+1, i} = \theta_{t, i} - \tfrac{\eta}{\sqrt{\mathcal{E}^{g \circ g}_{t+1, i} + \varepsilon}} \cdot m_{t, i}$
        \EndFor
    \end{algorithmic}
\end{algorithm}
\end{minipage}
\hspace{0.1cm}
\begin{minipage}{0.32\textwidth}
\begin{figure}[H]
  \centering
  \includegraphics[width=0.9\textwidth]{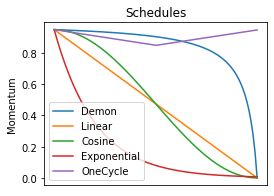}
  \caption{\emph{Non-linear} DEMON schedule vs. and other schedules, from $\beta_{init} = 0.9$ to 0.}
  \label{fig:demon_curve}
\end{figure}
\end{minipage}
\end{figure*}

\medskip
\noindent \textbf{Momentum tuning.} 
In this work, we focus on improving model performance and hyperparameter robustness with simple techniques for the momentum parameter.
Momentum was designed to speed up learning in directions of low curvature, without becoming unstable in directions of high curvature. 
To minimize the objective function $\mathcal{L}(\cdot)$, the most common momentum method, SGDM, is given by the following recursion: \vspace{-0.1cm}
\begin{align*}
\theta_{t+1} = \theta_{t} + \eta v_t, \quad v_{t} = \beta v_{t - 1} - g_t.
\end{align*}
 for variable $\theta_t \in \mathbb{R}^p$, where $\beta$ is the momentum, $g_t$ is a stochastic gradient, where $\mathbb{E}[g_t] = \nabla \mathcal{L}(\theta_t)$, and $\eta > 0$ is the step size.

Practitioners set $\beta = 0.9$.
This is supported by recent works \cite{chen2016revisiting, kingma2014adam, hinton2012neural, reddi2019convergence}, and by the fact that most common softwares, such as PyTorch \cite{paszke2017automatic}, use $\beta = 0.9$ as the default momentum value.
\emph{There is no indication that this choice is universally well-behaved.}

There are papers that attempt to tune the momentum parameter. 
In the distributed setting, \cite{mitliagkas2016asynchrony} observe that running SGD asynchronously is similar to adding a momentum-like term to SGD. They provide empirical evidence that setting $\beta = 0.9$ results in a momentum ``overdose'', yielding suboptimal performance.
YellowFin \cite{zhang2017yellowfin} is a learning rate and momentum adaptive method for both synchronous and asynchronous settings, motivated by a quadratic model analysis and some robustness insights. 
Finally, in training generative adversarial networks (GANs), optimal momentum values tend to decrease from $\beta = 0.9$ \cite{mirza2014conditional,radford2015unsupervised,arjovsky2017wasserstein}, taking even negative values \cite{gidel2018negative}.

\medskip
\noindent \textbf{This paper.} We perform the first large-scale empirical analysis of momentum decay methods and introduce the \textsc{Demon} momentum decay rule, a novel method which performs favorably and increases hyperparameter robustness in comparison to other learning rate and momentum schedules.
Our findings can be summarized as follows: \vspace{-0.1cm}
\begin{itemize}[leftmargin=*]
  \item We propose a new momentum decay rule, dubbed as \textsc{Demon}. \textsc{Demon} is motivated by decaying the total contribution of a gradient to all future updates, with limited additional computation. 
  \item Across 28 relevant settings, \textsc{Demon} achieves the highest ratio of Top-1 and Top-3 finishes: 39\% and 85\%, respectively. See Table \ref{tablepctresults}. These ratios nearly double those of the second-placed cosine learning rate decay schedule, which achieves 17\% and 60\%, respectively.
  In addition to outperforming other popular schedule such as the learning rate step schedule, linear schedule, and OneCycle, \textsc{Demon} also outperforms all other momentum schedules that were considered. 
  \item We observe improved robustness to hyperparameter tuning for \textsc{Demon} relative to the popular learning rate step schedule.
\end{itemize}
Experiments are provided on various datasets---including MNIST, FMNIST, CIFAR-10, CIFAR-100, STL-10, Tiny ImageNet, Penn Treebank (PTB), GLUE benchmark \cite{wang2019glue}; and networks---including Convolutional Networks (CNN) with Residual architecture (ResNet) \cite{he2016deep} (Wide ResNet) \cite{zagoruyko2016wide}, Non-Residual architecture (VGG-16) \cite{simonyan2014very}, Recurrent Neural Networks (RNN) with Long Short-Term Memory architecture (LSTM) \cite{hochreiter1997long}, Variational AutoEncoders (VAE) \cite{kingma2015vae}, Capsule Network \cite{sabour2017caps}, Noise Conditional Score Network (NCSN) \cite{song2019generative}, and BERT \cite{devlin2019bert}.

\section{Preliminaries} \label{preliminaries}


\noindent \textbf{SGDM.}
Let $\theta_t \in \mathbb{R}^p$ be the parameters of the network at time step $t$, where $\eta \in \mathbb{R}$ is the learning rate, and $g_t$ is the stochastic gradient w.r.t. $\theta_t$ for empirical loss $\mathcal{L}(\cdot)$. 
SGDM is parameterized by $\beta \in \mathbb{R}$, the momentum coefficient, and follows the recursion:
\begin{align*}
\theta_{t+1} = \theta_t + \eta v_{t}, \quad v_{t} = \beta v_{t-1} - g_t,
\end{align*}
where $v_t \in \mathbb{R}^p$ accumulates momentum.
Observe that for $\beta = 0$, the above recursion is equivalent to SGD. 
Common values for $\beta$ are closer to one, with $\beta = 0.9$ the most used value \cite{ruder2016overview}.

\medskip
\noindent \textbf{Adaptive gradient descent methods.}
These algorithms utilize current and past gradients to design preconditioning matrices that better approximate the local curvature of  $\mathcal{L}(\cdot)$ \cite{duchi2011adaptive, hinton2012neural}.
Adam \cite{kingma2014adam} uses a decaying average of past gradients, as in $\mathcal{E}^{g}_{t+1} = \beta_1 \cdot \mathcal{E}^{g}_t + (1-\beta_1) \cdot g_t$, as well as a decaying average of squared gradients, as in $\mathcal{E}^{g \circ g}_{t+1} = \beta_2 \cdot \mathcal{E}^{g \circ g}_t + (1 - \beta_2) \cdot (g_t \circ g_t)$, 
leading to the recursion:\footnote{For clarity, we will skip the bias correction step in this description of Adam; see \cite{kingma2014adam}.} 
\begin{align*}
\theta_{t+1, i} &= \theta_{t, i} - \tfrac{\eta}{\sqrt{\mathcal{E}^{g \circ g}_{t+1, i} + \varepsilon}} \cdot \mathcal{E}^g_{t+1, i}, \quad \forall t, 
\end{align*}
where usually $\beta_1 = 0.9$ and $\beta_2 = 0.999$.
\section{\textsc{Demon}: Decaying momentum algorithm}


\textsc{Demon} is motivated by learning rate decay models which reduce the impact of a gradient to current and future updates. 
By decaying momentum, we decay the total contribution of a gradient to all future updates. 
\emph{Our goal here is to present a concrete, effective, and easy-to-use momentum decay procedure, which we show in the experimental section.}
The key component is the momentum decay schedule: 
\begin{align} \label{momDecayRule}
    \beta_t = \beta_{\text{init}} \cdot \tfrac{ \left(1 - \frac{t}{T} \right) }{(1 - \beta_{\text{init}}) + \beta_{\text{init}} \left(1 - \frac{t}{T}\right)} \: . \\[-14pt] \nonumber
\end{align}
Above, the fraction $(1 - t/T)$ refers to the proportion of iterations remaining.
Fig. \ref{fig:demon_curve} presents a visualization of the proposed momentum decay rule and other common schedules. 
The interpretation of this rule comes from the following argument: Assume fixed momentum parameter $\beta_t \equiv \beta$; e.g., $\beta = 0.9$, as literature dictates.
For our discussion, we will use the SGDM recursion. 
We know that $v_0 = 0$, and $v_t = \beta v_{t-1} -  g_{t}$.
Then, the main recursion can be unrolled into: 
\begin{align*} 
\theta_{t+1}
             &= \theta_t - \eta g_t - \eta \beta g_{t-1} - \eta \beta^2 g_{t-2} + \eta \beta^3 v_{t-2} \\
             &= \cdots = \theta_{t} - \eta {g}_{t} - \eta \cdot \sum_{i = 1}^{t} \left(\beta^{i} \cdot {g}_{t - i}\right) \\[-16pt] \nonumber
\end{align*}

Interpreting the above recursion, 
\emph{a particular gradient term $g_{t}$ contributes a total of $\eta \sum_{i} \beta^{i}$ of its ``energy'' to all future gradient updates.} 
Moreover, for an asymptotically large number of iterations, we know that $\beta$ contributes on up to $t-1$ terms. Then, $\sum_{i = 1}^{\infty} \beta^{i} = \beta \sum_{i = 0}^{\infty} \beta^{i} = \beta/(1 - \beta)$.
Thus, in our quest for a simple momentum decay schedule, it is natural to consider a scheme where the cumulative momentum is decayed to $0$. 
Let $\beta_{\text{init}}$ be the initial $\beta$; then at the current step $t$ with a total of $T$ steps, we design the decay routine such that: ${\beta}/(1 - \beta) = (1 - {t}/{T}) \beta_{\text{init}}/(1 - \beta_{\text{init}})$.
This leads to \eqref{momDecayRule}. Although $\beta$ changes in subsequent iterations, this is typically a very close approximation because $\beta^i\beta^{i+1}\dots\beta^{t}$ for a particular $g^{i}$ diminishes much faster than $\beta$ changes. 

\medskip
\noindent \textbf{Formal intuition.}
Conceptually, \texttt{Demon} takes advantage of momentum for speed-up in the early phases of training, but decays momentum throughout training. 
This prevents neural network weights from growing too quickly.
Constraining weights has been recently studied and shown to be key for optimal performance.
In particular, it is demonstrated theoretically and empirically in \cite{heo2020adamp} that plain use of momentum leads to suboptimal performance. 
It is important to stabilize the growth of the weights across training. 

Formally, a function $f$ is scale invariant if $f(cx) = f(x)$ for constant $c > 0$. 
Let the function $\texttt{Norm}(\cdot)$ be defined as $\texttt{Norm}_s (x) = (x - \mu_s (x)) / \sigma_s(x)$, where $s$ is a subset of the dimensions of $x$, $\mu_s$ is the mean for those dimensions, and $\sigma_s$ the standard deviation. 
$\texttt{Norm}(\cdot)$ is scale invariant, and includes cases such as Batch Norm \cite{heo2020adamp}. 
Concretely, $\texttt{Norm}(\theta^{T} x) = \texttt{Norm}((c\theta)^{T} x)$. 
With adaptive $\beta_t$ on iteration $t$, we extend the norm growth lemma in \cite{heo2020adamp} as follows:
\begin{lemma} \label{lemma:0} 
    \text{For scale invariant $\theta$ given by momentum SGD, }\newline\text{the following equality holds true:}
    \begin{align*}
    {\norm{\theta_{t + 1}}}_{2}^{2} = {\norm{\theta_{t}}}_{2}^{2} + \eta^2 {\norm{v_t}}_{2}^{2} + 2\eta^2 \sum_{i=0}^{t - 1} \beta_{i} \beta_{i + 1} \dots \beta_{t - 1} {\norm{v_i}}_{2}^{2}
    \end{align*}
\end{lemma}

Since the latter term grows increasingly larger, \texttt{Demon} decays the momentum to curb its growth, and therefore reduces the norm growth of the parameters, leading to improved performance \cite{heo2020adamp}. The extension is trivial and the proof is omitted.


\medskip
\noindent \textbf{Connection to previous algorithms.}
\textsc{Demon} introduces an implicit discount factor.
The main recursions of the algorithm resemble those of standard machine learning algorithms.
E.g., for $\beta_t = \beta = 0.9$ we obtain SGD with momentum, and for $\beta = 0$ we obtain plain SGD in Algorithm \ref{alg:demon_sgdm}; in Algorithm \ref{alg:demon_adam}, for $\beta_1 = 0.9$ with a slight adjustment of learning rate we obtain Adam, while for $\beta_1 = 0$ we obtain a non-accumulative AdaGrad algorithm. 
We choose to apply \textsc{Demon} to a slightly adjusted Adam to isolate the effect of the momentum parameter, since the momentum parameter adjusts the magnitude of the current gradient as well in vanilla Adam.







\medskip
\noindent \textbf{Efficiency.}
\textsc{Demon} requires only limited extra overhead and computation in comparison to the vanilla counterparts for the computation of $\beta_t$. Implementation is simply 1-2 lines of code. 
\begin{small} 
    \begin{align*}
        &\rm\texttt{p\_t} = ({\rm\texttt{iters - t}}) ~/~ {\rm\texttt{iters}} \\
        &\rm\texttt{beta\_t} = \rm\texttt{beta1} ~*~ (\rm\texttt{p\_t} ~/~ (1 - \rm\texttt{beta1} + \rm\texttt{beta1} * \rm\texttt{p\_t}))  \\[-16pt]
    \end{align*}
    \end{small}

\medskip\noindent \textbf{Convergence analysis.}
\emph{We provide convergence proof for \textsc{Demon} SGDM in the convex setting, by bounding auxiliary function sequences} \cite{ghadimi2014global}.
For an objective function $f$ which is convex, continuously differentiable, its gradient $\nabla f(\cdot)$ is Lipschitz continuous with constant $L$, our goal is to show that $f(\bar{\theta}_T)$ converges to the optimum $f^*$ with decreasing momentum, where $\bar{\theta}_T$ is the average of $\theta_t$ for $t=1,...,T$, following \cite{ghadimi2014global}. Our following theorem holds for a constant learning rate and $\beta_t$ decaying with $t$.
\begin{theorem} \label{convergenceTheorem}
Assume that $f$ is convex, continuously differentiable, its gradient $\nabla f(\cdot)$ is Lipschitz continuous with constant $L$, with a decreasing momentum, but constant step size, as in:
    $\beta_t = \tfrac{1}{t} \cdot \tfrac{t + 1}{t + 2}, \alpha \in \left( 0, \tfrac{2}{3L} \right)$.
We consider the SGDM iteration in non-stochastic settings, where:
    $\theta_{t+1} = \theta_t - \alpha \nabla f(\theta_t) + \beta_t \left(\theta_{t} - \theta_{t-1}\right)$.
Then, the sequence $\{\theta_t\}_{t = 1}^T$ generated by the SGDM iteration, with decreasing momentum, satisfies:
\begin{align*}
    f(\bar{\theta}_T) - f^* \leq \tfrac{\Vert{\theta_1 - \theta^\star} \Vert^2}{T} \left( \tfrac{3}{4} L + \tfrac{1}{2 \alpha} \right),
\end{align*}
where $\bar{\theta}_T$ is the Cesaro average of the iterates: $\bar{\theta}_T = \tfrac{1}{T} \sum_{t=1}^{T} \theta_t$.
\end{theorem}

\textit{Proof.}
Let $\beta_t = \frac{1}{t} \cdot \frac{t + 1}{t + 2}$ and 
    $p_t = \tfrac{1}{t} (\theta_t - \theta_{t - 1})$. 
We consider the SGDM iteration in non-stochastic settings, where
    $\theta_{t+1} = \theta_t - \alpha \nabla f(\theta_t) + \beta_t \left(\theta_{t} - \theta_{t-1}\right)$.
Using the definition of $p_t$ above, one can easily prove that:
\begin{equation*}
    \theta_{t + 1} + p_{t + 1} = \left(1 + \frac{1}{t + 1}\right)\theta_{t + 1} - \frac{1}{t + 1}\theta_t = \theta_t + p_t - \tfrac{\alpha(t + 2)}{t + 1} \nabla f(\theta_t).
\end{equation*}
Using this expression, we will analyze the term $\|\theta_{t+1} + p_{t+1} - \theta^\star\|_2$:
\begin{align*}
    \Vert \theta_{t + 1} + p_{t + 1} - \theta^\star \Vert^2 &= \Vert \theta_{t} + p_{t} - \theta^\star \Vert^2 \\
    & \quad \quad - \tfrac{2 \alpha (t + 2)}{t + 1} \left \langle \theta_{t} + p_{t} - \theta^\star , \nabla f(\theta_t) \right \rangle \\
    & \quad \quad \quad \quad+ \left( \tfrac{\alpha(t + 2)}{t + 1} \right)^2 \cdot \Vert \nabla f(\theta_t) \Vert^2 \\
    &= \Vert \theta_{t} + p_{t} - \theta^\star \Vert^2 - \tfrac{2 \alpha (t + 2)}{t(t + 1)} \left \langle \theta_{t} - \theta_{t - 1} , \nabla f(\theta_t) \right \rangle \\
    & \quad \quad - \tfrac{2 \alpha (t + 2)}{t + 1} \left \langle \theta_{t} - \theta^\star , \nabla f(\theta_t) \right \rangle  \\
    & \quad \quad \quad \quad + \left( \tfrac{\alpha (t + 2)}{t + 1}\right)^2 \cdot \Vert \nabla f(\theta_t) \Vert^2
\end{align*}
Since $f$ is convex, continuously differentiable, its gradient is Lipschitz continuous with constant L, then
\begin{align}\label{equation_assump1}
    \tfrac{1}{L} \Vert \nabla f(\theta_t) \Vert^2 & \leq \langle \theta_t - \theta^\star, \nabla f(\theta_t) \rangle, \\ \label{equation_assump2}
    f(\theta_t) - f^* + \tfrac{1}{2L} \Vert \nabla f(\theta_t) \Vert^2 & \leq \langle \theta_t - \theta^\star, \nabla f(\theta_t) \rangle, \\
    f(\theta_t) - f(\theta_{t - 1})  & \leq \langle \theta_t - \theta_{t - 1}, \nabla f(\theta_t) \rangle.
\end{align}
Substituting the above inequalities leads to
\begin{align*}
    \Vert \theta_{t + 1} + p_{t + 1} - \theta^\star \Vert^2 &\leq \Vert \theta_{t} + p_{t} - \theta^\star \Vert^2 - \tfrac{2 \alpha (t + 2)}{t(t + 1)} (f(\theta_{t}) - f(\theta_{t - 1}))  \\
    & \quad \quad - 2 \alpha \tfrac{(1 - \lambda)(t + 2)}{L(t + 1)} \cdot \Vert \nabla f(\theta_t) \Vert^2 \\
    & \quad \quad \quad - 2 \alpha \lambda \tfrac{t + 2}{t + 1} (f(\theta_t) - f^*) \\
    & \quad \quad \quad \quad - \left( \alpha \tfrac{\lambda (t + 2)}{L(t + 1)}\right) \cdot \Vert \nabla f(\theta_t) \Vert^2 \\
    & \quad \quad \quad \quad \quad + \left( \frac{\alpha (t + 2)}{t + 1}\right)^2 \cdot \Vert \nabla f(\theta_t) \Vert^2
\end{align*}
where $\lambda \in (0, 1]$ is a parameter weighting (\ref{equation_assump1}) and (\ref{equation_assump2}). Grouping together terms yields
\begin{align*}
    &\left(\tfrac{2\alpha (t + 2)}{t(t + 1)} + \tfrac{2\alpha\lambda(t + 2)}{t + 1} \right) (f(\theta_t) - f^* ) + \Vert \theta_{t + 1} + p_{t + 1} - \theta^\star \Vert^2
    \leq \\
    &\quad \quad \tfrac{2\alpha(t + 2)}{t(t + 1)} (f(\theta_{t - 1}) - f^*) + \Vert \theta_{t} + p_{t} - \theta^\star \Vert^2 \\
    &\quad \quad \quad \quad + \tfrac{ \alpha(t + 2)}{t + 1} \left( \tfrac{\alpha (t + 2)}{t + 1} - \tfrac{2(1 - \lambda)}{L} - \tfrac{\lambda}{L} \right) \Vert \nabla f(\theta_t) \Vert^2 .
\end{align*}
The last term is non-positive when $\alpha \in [0, \frac{t + 1}{t + 2}(\frac{2 - \lambda}{L})]$ so it can be dropped. 
Summing over $t=1,...,T$ yields
\begin{align*}
    &2\alpha\lambda \sum_{t=1}^{T} \tfrac{t + 2}{t + 1}(f(\theta_t) - f^*) \\
    &+ \sum_{t = 1}^{T} \left(  \tfrac{2\alpha(t + 2)}{t(t + 1)}(f(\theta_t) - f^*) + \Vert \theta_{t + 1} + p_{t + 1} - \theta^\star \Vert^2 \right) \leq \\
    &\sum_{t = 1}^{T} \left(\tfrac{2\alpha(t + 2)}{t(t + 1)}(f(\theta_{t- 1}) - f^*) + \Vert \theta_{t} + p_{t} - \theta^\star \Vert^2 \right),
\end{align*}
implying that:
\begin{align*}
    2\alpha\lambda \sum_{t = 1}^{T} \tfrac{t + 2}{t + 1} (f(\theta_t) - f^*) \leq 3\alpha (f(\theta_1) - f^*) + \Vert \theta_1 - \theta^\star \Vert^2.
\end{align*}
Since: $2\alpha\lambda \sum_{t = 1}^{T} (f(\theta_t) - f^*) \leq 2\alpha\lambda \sum_{t = 1}^{T} \tfrac{t + 2}{t + 1} (f(\theta_t) - f^*)$ \newline$\leq 3\alpha\lambda \sum_{t = 1}^{T}  (f(\theta_t) - f^*)$,
we further have:
\begin{align*}
    3\alpha\lambda \sum_{t = 1}^{T}  (f(\theta_t) - f^*) \leq \tfrac{3}{2} \bigg( 3 \alpha (f(\theta_1) - f^*) + \Vert \theta_1 - \theta^\star \Vert^2 \bigg).
\end{align*}
Due to the convexity of $f$:
    $f(\bar{\theta}_t) \leq \tfrac{1}{T} \sum_{t=1}^{T} f(\theta_t)$,
observe that
\begin{align*}
    &f(\bar{\theta}_T) - f^* \leq \tfrac{1}{T} \sum_{t = 1}^{T}  (f(\theta_t) - f^*) \leq \\ &\tfrac{1}{3\alpha\lambda T} \left(\tfrac{9}{2}\alpha (f(\theta_1) - f^*) + \tfrac{3}{2}\Vert \theta_1 - \theta^\star \Vert^2 \right).
\end{align*}
Since $f(\theta_1) - f^* \leq \tfrac{L}{2} \Vert \theta_1 - \theta^\star \Vert^2$ by Lipschitz continuous gradients, setting $\lambda = 1$ and observing $(t + 1) / (t + 2) \geq 2/3$ gives the result.

For \textsc{Demon} Adam, we observe it lies within the definition of Generic Adam in \cite{zou2018a}, and inherits the non-convex results. Namely, to restate the theorem \cite{zou2018a}:

\begin{theorem}
Assuming $\tau \in \{1, 2, \dots, T\}$ is an index over $T$ iterations, Demon + Adam satisfies: \vspace{-0.5cm}
\end{theorem}
\begin{align*}
\mathbb{E}[ \|\nabla f(x_{\tau})\|_2^{4/3} ]^{3/2}  \leq \frac{C + C' \sum_t \alpha_t \sqrt\{1 - \theta_t\}}{\alpha_T \cdot T} = O\left(\frac{1}{T}\right),
\end{align*}

where $C, C'$ are constants, $\alpha_t$ are step sizes, $\theta_t$ are the parameters related to the diagonal Hessian approximation of Adam's preconditioner, and $\beta_t < 1$, as is the case for Demon technique.

Moreover, one can remove the expectation requirement above, and have the same result hold deterministically with some probability $1 - \delta^{2/3}$:
\begin{align*}
\|\nabla f(x_{\tau})\|_2^2 \leq \frac{C + C' \sum_t \alpha_t \sqrt\{1 - \theta_t\}}{\delta \cdot \alpha_T \cdot T} = O\left(\frac{1}{T}\right).
\end{align*}

To achieve the above, the assumptions are lower-bounded function $f$, $L$-smoothness, and standard assumptions on gradients. Regarding the parameters, momentum $\beta_t$ has to satisfy $0 \leq \beta_t \leq \beta < 1$, for some $\beta$: this is exactly the setting of Demon, where $\beta$ is the initial value of the momentum, and $\beta_t$ decreases to zero. Our work is an exact instance of decreasing momentum that leads to empirical improvements compared to previous work; to the best of our knowledge, no other paper has considered a specific decreasing schedule for momentum that is at the same time almost "hyperparameter-free".



\medskip
\noindent \textbf{Practical suggestions.}
We advocate for decaying momentum from $\beta_{\text{init}}$ at $t=0$, to $0$ at $t=T$ as a general rule, which is the setting we use for all \textsc{Demon} experiments in this paper.

\section{Related work}

Numerous techniques exist for automatic hyperparameter tuning. Adaptive methods, such as AdaGrad \cite{duchi2011adaptive}, AdaDelta \cite{zeiler2012adadelta}, RMSprop \cite{hinton2012neural}, and Adam \cite{kingma2014adam}, are most widely used. 
Interest in closing the generalization difference between adaptive methods and SGDM led to AMSGrad \cite{reddi2019convergence}, which uses the maximum of the exponential moving average of squared gradients, QHAdam \cite{ma2018quasi}, a variant of QHM that recovers a variety of optimization algorithms, AdamW \cite{loshchilov2017fixing}, which decouples weight decay in Adam, and Padam \cite{chen2018closing}, which lowers the exponent of the second moment. 
YellowFin \cite{zhang2017yellowfin} is a learning rate and momentum adaptive method motivated by a quadratic model analysis and robustness insights. In the non-convex setting, STORM \cite{cutkosky2019momentum} uses a variant of momentum for variance reduction.

The convergence of momentum methods has been heavily explored both empirically and theoretically \cite{wibisono2015accelerated,wibisono2016variational,wilson2016lyapunov,kidambi2018insufficiency,defazio2020factorial}. 
\cite{sutskever2013importance} explored momentum schedules, even increasing momentum during training, inspired by Nesterov's routines for convex optimization. \cite{smith2017increasebatch} scales the batch size to create associated changes in learning rate and momentum. \cite{smith20181cycle} introduces cycles of simultaneously increasing learning rate and decreasing momentum followed by simultaneously decreasing learning rate and increasing momentum.
Some work adapts the momentum to reduce oscillations during training \cite{o2015adaptive} and explores integration of momentum into well-conditioned convex problems \cite{srinivasan2018adine}. Another approach is to combine several momentum vectors with different $\beta$ values \cite{lucas2018aggregated}. In another work, gradient staleness in variance reduction methods is addressed with gradient transport \cite{arnold2019reducing}. We are aware of the theoretical work of \cite{yuan2016sgdequivalence} that proves, under certain conditions, SGDM is equivalent to SGD with a rescaled learning rate, but our experiments in the deep learning setting show slightly different behavior. Understanding this discrepancy is an exciting direction of research.

Smaller values of $\beta$ have been employed for Generative Adversarial Networks (GANs), and recent developments in game dynamics \cite{gidel2018negative} show a negative momentum is helpful.
\section{Experiments} \label{experiments}
Well-known experiments in the literature are selected for comparison (e.g., ResNets, LSTMs, and BERT). 
\emph{Training models like GPT-2/3 from scratch is not feasible, and we instead focus on providing a wide number of experiments and baselines.}  
For each setting, we use varying numbers of epochs to demonstrate effectiveness. 
Experiments with different numbers of epochs are standalone experiments with independently-tuned hyperparameters. All settings are summarized in Table \ref{experimentSummary} and comprehensively detailed in Appendix \ref{detailedExperimentSettings}. 

\begin{table}[!htp]
\centering
\begin{small}
\caption{\small Summary of experimental settings.}\label{experimentSummary}
\begin{tabular}{l|l|l} 
\toprule
    Experiment short name & Model & Dataset \\ \midrule
    \texttt{RN20-CIFAR10}  & ResNet20  & CIFAR10  \\
    \texttt{RN56-TINYIMAGENET} & ResNet56 & Tiny ImageNet \\
    \texttt{VGG16-CIFAR100} & VGG-16 & CIFAR100  \\
    \texttt{WRN-STL10} & Wide ResNet 16-8 & STL10  \\
    \texttt{LSTM-PTB} & LSTM RNN & Penn TreeBank  \\
    \texttt{VAE-MNIST} & VAE & MNIST  \\
    \texttt{NCSN-CIFAR10} & NCSN & CIFAR10  \\
    \texttt{CAPS-FMNIST} & Capsnet & FMNIST \\
    \texttt{BERT$_{BASE}$-GLUE} & BERT (Pre-trained) & GLUE (9 tasks) \\
 \bottomrule
\end{tabular}
\end{small}
\end{table}


\begin{figure*}[t]
  \centering
  \caption{\small
  Left to right: Error rate for \texttt{WRN-STL10-DEMONSGDM} (top) and \texttt{WRN-STL10-SGDM} (bottom) for 50 epochs, error rate for \texttt{VGG16-CIFAR100-DEMONSGDM} (top) and \texttt{VGG16-CIFAR100-SGDM} (bottom) for 100 epochs, and error rate for \texttt{RN20-CIFAR10-DEMONAdam} (top) and \texttt{RN20-CIFAR10-Adam} (bottom) for 100 epochs. Light-colored patches indicate better performance.}
  \begin{subfigure}[b]{0.33\linewidth}
    \includegraphics[width=\linewidth]{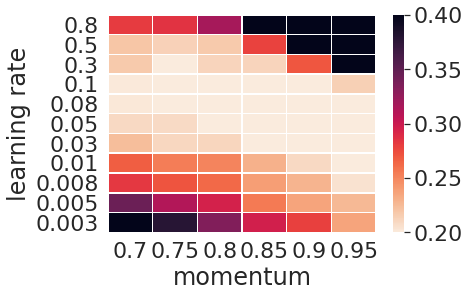}
    \label{fig:stl10demonsgdmheatmap}
  \end{subfigure}
  \hspace{0.1cm}
  \begin{subfigure}[b]{0.33\linewidth}
    \includegraphics[width=\linewidth]{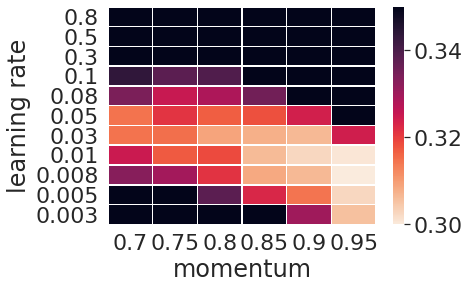}
    \label{fig:stl10demonsgdmheatmap}
  \end{subfigure}
  \hspace{0.1cm}
  \begin{subfigure}[b]{0.30\linewidth}
    \includegraphics[width=\linewidth]{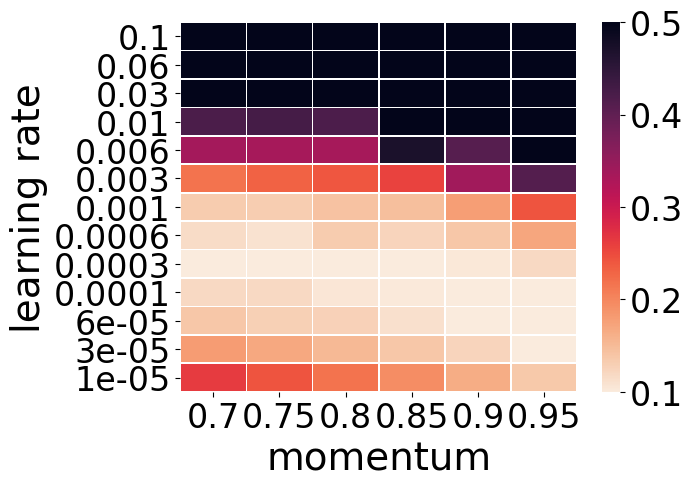}
    \label{fig:cifar10demonadamheatmap}
  \end{subfigure}
  \newline
  \centering
    \begin{subfigure}[b]{0.34\linewidth}
    \includegraphics[width=\linewidth]{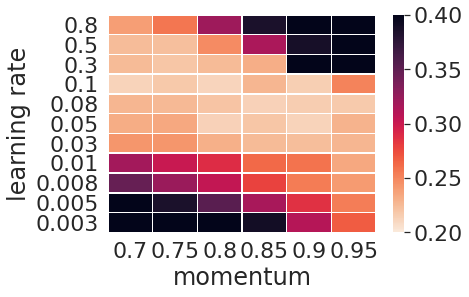}
    \label{fig:stl10sgdmheatmap}
  \end{subfigure}
  \begin{subfigure}[b]{0.34\linewidth}
    \includegraphics[width=\linewidth]{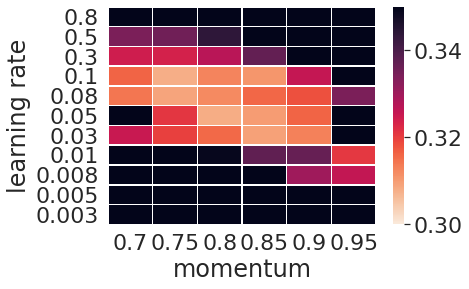}
    \label{fig:stl10sgdmheatmap}
  \end{subfigure}
    \begin{subfigure}[b]{0.30\linewidth}
    \includegraphics[width=\linewidth]{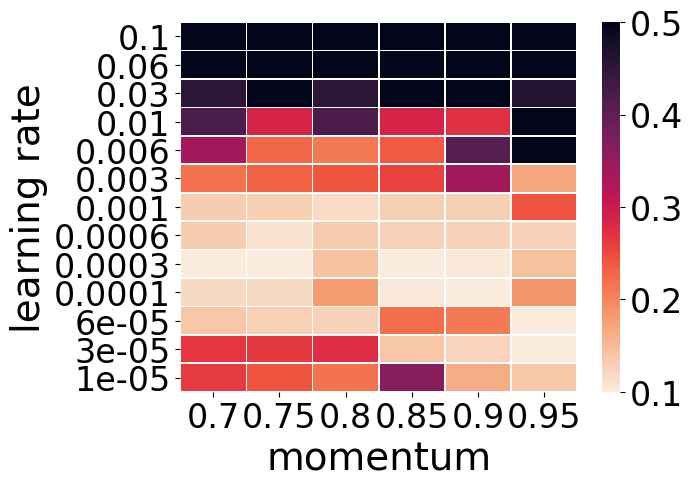}
    \label{fig:cifar10adamheatmap}
  \end{subfigure}
  \label{fig:heatmaps}
\end{figure*}

We note that we have implemented setups that use the standard ImageNet dataset as input.
However, we did not observe any results worth noting between learning rate step decay and \textsc{Demon}. 
In particular, using a ResNet-50, comparable performance to state of the art is achieved using most algorithms under consideration. 
ImageNet results are often not indicative of generalizability, see \cite{pmlr-v119-shankar20c}, and this highlights that a breadth of experiments is more important.

We consider the following schedules, where we tune both the learning rate in multiples of 3, the momentum $\in$ [0.9,0.95,0.97], and weight decay, in addition to the other hyperparameters specialized to the particular schedule. The cost refers to if there are any other hyperparameters to tune that are specialized to the schedule.
\begin{itemize}[leftmargin=*]
    \item \emph{No schedule:} Self-explanatory (\textbf{1$\times$ cost}).
    \item \emph{Step schedule:} One of the most common kind of schedules (at this moment) for achieving state-of-the-art results in the literature \cite{hu2017squeeze,huang2017densely, lin2017feature,   wang2017residual,zagoruyko2016wide}. We attempt decay schedules including 0.1$\times$ at 50\% and 75\% of total epochs; 0.1$\times$ at 25\%, 50\%, 75\%; 0.1$\times$ at 33\% and 66\%; 0.1$\times$ at 10\%, 25\%, 50\%, 75\% (\textbf{4$\times$ cost}).
    \item \emph{Cosine schedule\cite{loshchilov2017sgdr}:} Follows the smooth schedule of $\gamma_t = \gamma_{min} + 0.5 \cdot (\gamma_{max} - \gamma_{min}) (1 + cos(\pi t / T))$. $\gamma$ can be the learning rate or the momentum. We consider $\gamma_{min} = 0$ to achieve (\textbf{1$\times$ cost}). 
    \item \emph{OneCycle\cite{smith20181cycle}:} This scheme roughly follows increasing learning rate linearly from $0.1\eta$ to $\eta$ and back to $0.1 \eta$, while simultaneously decreasing momentum linearly from $\beta_{max}$ to $\beta_{min}$ and back to $\beta_{max}$. For the momentum values, we consider the pairs [(0.95, 0.85), (0.9, 0.85), (0.95, 0.9)]. For the momentum variant, we simply consider the momentum component of OneCycle (\textbf{1$\times$ cost}).
    \item \emph{Linear schedule:} Decreases the hyperparameter from the initial value to 0 across epochs (\textbf{1$\times$ cost}).
    \item \emph{Exponential schedule\cite{45381}:} Follows the smooth schedule of $\gamma_t = \gamma \cdot e^{kt}$. We tune $k$ from a reasonable starting point, $-0.05 \cdot (100/T)$ which is a scaled version of the curve in Figure \ref{fig:demon_curve} (i.e., plotted for 100 iterations in multiplies of 2) (\textbf{$\sim$4$\times$ cost}). 
    \item \emph{Decay on Plateau\cite{45381}:} Commonly used in practice where the learning rate is multiplied by a factor if validation loss stops improving for a certain number of epochs (patience). We tune the patience in multiples of 5, and multiply the learning rate by 0.1 (\textbf{$\sim$5$\times$ cost}).  
    \item \textsc{Demon}: This paper. The schedule follows Algorithm \ref{momDecayRule} and decays to 0 at the end of training (\textbf{1$\times$ cost}).
\end{itemize}
We apply these schedules to SGDM and Adam, focusing on the performance of different schedules. \emph{The overall tuning budget for \textsc{Demon} SGDM/Adam is generally equal to or less than that of SGDM/Adam with its relevant possible schedules}.

\subsection{Decreasing the need for tuning} \label{decreasingTheNeed}
We demonstrate the hyperparameter robustness of \textsc{Demon} SGDM and \textsc{Demon} Adam relative to SGDM with learning rate step schedule and Adam. In Fig. \ref{fig:heatmaps}, validation accuracy is displayed for a grid search over learning rate and momentum. For SGDM, results are obtained with the highest-performing learning rate schedule. 
The heatmaps display optimal performance of each optimizer over the full range of possible hyperparameters. 

\texttt{WRN-STL10-DEMONSGDM} yields a significantly larger band of lighter color, indicating better performance for a wide range of hyperparameters. For every learning rate-momentum pair, we observe a lighter color for \textsc{Demon} SGDM relative to SGDM. Concretely, SGDM has roughly one configuration per column with $<22\%$ generalization error, while \textsc{Demon} SGDM has five. 

On \texttt{VGG16-CIFAR100-DEMONSGDM}, a larger band of low generalization error exists compared to SGDM. There also appears to be a slight shift in optimal parameters. Concretely, \textsc{Demon} SGDM has almost three times the number of configurations with generalization error $<31\%$. 

On \texttt{RN20-CIFAR10}, \textsc{Demon} Adam demonstrates its improved hyperparameter robustness relative to Adam. The generalization errors achieved with Adam fluctuate significantly, yielding optimal performance with only a few hyperparameter settings. In contrast, \textsc{Demon} Adam yields a wide band of high performance across hyperparameters.

\begin{table*}[t]
\centering
\begin{small}
\caption{ \texttt{RN20-CIFAR10}, \texttt{WRN-STL10} and \texttt{RN56-TINYIMAGENET} generalization error. The number of epochs was predefined before the execution of the algorithms. Red indicates Top-1 performance, bold is Top-3, ignoring non SGDM and Adam optimizers.}
\begin{tabular}{c|ccc|ccc|cc}
\toprule
    & \multicolumn{3}{c}{\textbf{ResNet 20}} & \multicolumn{3}{c}{\textbf{Wide ResNet 16-8}} & \multicolumn{2}{c}{\textbf{ResNet 56}} \\ \midrule
    SGDM & 75 epochs & 150 epochs  & 300 epochs & 50 epochs & 100 epochs & 300 epochs \\ \midrule
    + LR Step Schedule & \textbf{8.82} $\pm$ .25 & 8.43 $\pm$ .07 & \textcolor{myred}{\textbf{7.32}} $\pm$ .14 & 22.42 $\pm$ .56 & \textbf{17.20} $\pm$ .35 & \textbf{14.51} $\pm$ .26 & 45.98 $\pm$ .23 & 41.66 $\pm$ .10 \\
    + LR Cosine Schedule  & \textbf{8.80} $\pm$ .08 & \textcolor{myred}{\textbf{8.10}} $\pm$ .13 & 7.78 $\pm$ .14 & \textbf{20.03} $\pm$ .26 & \textbf{17.02} $\pm$ .24 & 14.66 $\pm$ .25 & - & -\\    
    + OneCycle  & 10.83 $\pm$ .25 & 9.23 $\pm$ .19 & 8.42 $\pm$ .12 & 21.67 $\pm$ .27 & 19.69 $\pm$ .21 & 19.00 $\pm$ .42  & - & -\\  
    + LR Linear Schedule  & 9.03 $\pm$ .24 & \textbf{8.15} $\pm$ .12 & \textbf{7.62} $\pm$ .12 & \textcolor{myred}{\textbf{19.54}} $\pm$ .20 & 17.39 $\pm$ .24 & \textbf{14.58} $\pm$ .18  & - & -\\ 
    + LR Decay on Plateau  & 9.05 $\pm$ .07 & 8.26 $\pm$ .07 & 7.97 $\pm$ .14 & 21.05 $\pm$ .27 & 17.83 $\pm$ .39 & 15.16 $\pm$ .36  & - & -\\ 
    + LR Exp decay  & 9.55 $\pm$ .09 & 9.20 $\pm$ .13 &  7.82 $\pm$ .05 & 22.65 $\pm$ .49 & 20.60 $\pm$ .21 & 15.85 $\pm$ .28  & - & -\\ 
    + OneCycle Momentum  & 15.61 $\pm$ .39 & 14.02 $\pm$ .13 & 13.35 $\pm$ .58  & 29.34 $\pm$ .78 & 23.20 $\pm$ .39 & 25.42 $\pm$ .47   & - & -\\ 
    + Cosine Momentum  & 13.57 $\pm$ .20 & 11.23 $\pm$ .22 & 10.87 $\pm$ .03 & 24.12 $\pm$ .34 & 21.66 $\pm$ .37 & 16.29 $\pm$ .26  & - & -\\ 
    + Linear Momentum  & 12.31 $\pm$ .14 & 10.26 $\pm$ .16 & 10.63 $\pm$ .31  & 25.13 $\pm$ .28 & 22.74 $\pm$ .78 & 17.92 $\pm$ .41   & - & -\\ 
    + Exp Momentum decay  & 14.96 $\pm$ .19 & 12.98 $\pm$ .15 & 12.35 $\pm$ .11 & 24.01 $\pm$ .20 & 19.35 $\pm$ .29  & 17.56 $\pm$ .21  & - & - \\ \midrule
    AggMo + LR Step & 8.71 $\pm$ .24 & 7.93 $\pm$ .15 & 7.62 $\pm$ .03 & 21.37 $\pm$ .32 & 17.15 $\pm$ .35 & 14.49 $\pm$ .26  & - & -\\
    QHM + LR Step & 8.72 $\pm$ .14 & 7.95 $\pm$ .17 & 7.67 $\pm$ .10 & 21.75 $\pm$ .31 & 18.21 $\pm$ .48 & 14.44 $\pm$ .23  & - & -\\ \midrule
    \textsc{Demon} SGDM & \textcolor{myred}{\textbf{8.56}} $\pm$ .10 & \textbf{8.21} $\pm$ .18 & \textbf{7.59} $\pm$ .12 & \textbf{20.23} $\pm$ .31 & \textcolor{myred}{\textbf{16.19}} $\pm$ .23 & \textcolor{myred}{\textbf{14.44}} $\pm$ .53  & 44.87 $\pm$ .15 & 40.85 $\pm$ .01 \\ \midrule \midrule
    Adam & 13.63 $\pm$ .22 & 11.90 $\pm$ .06 & 11.94 $\pm$ .06 &  23.35 $\pm$ .20 & \textbf{19.63} $\pm$ .26 & 18.65 $\pm$ .07  & 57.56 $\pm$ 1.50 & 50.89 $\pm$ .59 \\
    + LR Step Schedule & 10.47 $\pm$ .10 & \textcolor{myred}{\textbf{8.75}} $\pm$ .17 & \textbf{8.55} $\pm$ .05 & 23.85 $\pm$ .07 & \textbf{19.63} $\pm$ .33 & \textbf{18.29} $\pm$ .10  & - & - \\  
    + LR Cosine Schedule  & \textbf{9.56} $\pm$ .12 & 9.15 $\pm$ .12 & 8.93 $\pm$ .07  & 22.85 $\pm$ .47 & 21.47 $\pm$ .31 & 19.08 $\pm$ .36  & - & - \\    
    + OneCycle  & 10.33 $\pm$ .20 & 9.87 $\pm$ .12 & 9.03 $\pm$ .18 & \textcolor{myred}{\textbf{20.02}} $\pm$ .19 & \textcolor{myred}{\textbf{19.21}} $\pm$ .28 & 19.03 $\pm$ .43  & - & - \\  
    + LR Linear Schedule  & \textcolor{myred}{\textbf{9.25}} $\pm$ .12 & 9.20 $\pm$ .22 & 8.89 $\pm$ .05  & \textbf{21.70} $\pm$ .11 & 21.53 $\pm$ .44 & \textcolor{myred}{\textbf{17.85}} $\pm$ .15  & - & - \\ 
    + LR Decay on Plateau  & 9.71 $\pm$ .39 & \textbf{8.92} $\pm$ .18 & 8.80 $\pm$ .11 & 22.77 $\pm$ .33 & 19.91 $\pm$ .45 & 19.61 $\pm$ .56  & - & - \\ 
    + LR Exp decay  & 10.48 $\pm$ .15 & 9.24 $\pm$ .16 &  \textbf{8.53} $\pm$ .07 & 23.30 $\pm$ .39 & 20.70 $\pm$ .50 & 19.63 $\pm$ .24  & - & - \\ 
    + OneCycle Momentum  & 20.05 $\pm$ .91 & 15.60 $\pm$ .69 & 14.85 $\pm$ .50  & 24.61 $\pm$ .54 & 23.39 $\pm$ .39 & 23.54 $\pm$ .38  & - & - \\ 
    + Cosine Momentum  & 11.08 $\pm$ .11 & 10.63 $\pm$ .20 & 10.64 $\pm$ .30 & 25.76 $\pm$ .22 &  23.58 $\pm$ .02 & 20.10 $\pm$ .15  & - & - \\ 
    + Linear Momentum  & 11.91 $\pm$ .18 & 11.48 $\pm$ .13 & 11.09 $\pm$ .12  & 24.36 $\pm$ .31 & 21.93 $\pm$ .23 & 21.81 $\pm$ .36  & - & - \\  
    + Exp Momentum decay  & 15.18 $\pm$ .10 & 12.08 $\pm$ .16 & 10.63 $\pm$ .12 & 28.90 $\pm$ .21 & 25.28 $\pm$ .31 & 22.90 $\pm$ .41   & - & - \\ \midrule
    AMSGrad & 13.43 $\pm$ .14 & 11.83 $\pm$ .12 & 10.48 $\pm$ .12 &  21.73 $\pm$ .25 & 19.35 $\pm$ .20 & 18.21 $\pm$ .18  & - & - \\
    AdamW & 12.46 $\pm$ .52 & 11.38 $\pm$ .21 & 10.50 $\pm$ .17 & 20.39 $\pm$ .62 & 18.55 $\pm$ .23 & 17.00 $\pm$ .41  & - & -\\
    QHAdam & 15.55 $\pm$ .25 & 13.78 $\pm$ .08 & 13.36 $\pm$ .11 & 21.25 $\pm$ .22 & 19.81 $\pm$ .18 & 18.52 $\pm$ .25  & - & - \\
    YellowFin & 13.66 $\pm$ .34 & 12.13 $\pm$ .41 & 11.39 $\pm$ .16 & 22.55 $\pm$ .14 & 20.68 $\pm$ .04 & 18.56 $\pm$ .33  & - & -\\ \midrule
    \textsc{Demon} Adam & \textbf{9.68} $\pm$ .07 & \textbf{8.90} $\pm$ .18 & \textcolor{myred}{\textbf{8.50}} $\pm$ .12 & \textbf{20.95} $\pm$ .23 & \textbf{19.50} $\pm$ .32 & \textbf{18.62} $\pm$ .41  & 48.92 $\pm$ .03 & 45.72 $\pm$ .31 \\
 \bottomrule
\end{tabular}
\label{tableresnets}
\end{small}
\end{table*}

These results suggest that both \textsc{Demon} Adam and \textsc{Demon} SGDM are less sensitive to hyperparameter tuning than their learning rate step schedule counterparts, whilst attaining competitive error. This is critical to the use of \textsc{Demon} in practice, as \textsc{Demon} can yield high performance with minimal tuning. The performance of \textsc{Demon} is high and stable across a wide range of hyperparameters near the default.


\subsection{Results} \label{comparisonOfMethods}
For benchmarking purposes, we also include some other baselines where the learning rate and/or momentum are automatically adapted. These include Quasi Hyperbolic Adam (QHAdam) \cite{ma2018quasi}, Quasi Hyperbolic Momentum (QHM) \cite{ma2018quasi}, AMSGrad \cite{reddi2019convergence}, AdamW \cite{loshchilov2017fixing}, YellowFin \cite{zhang2017yellowfin}, Aggregated Momentum (AggMo) \cite{lucas2018aggregated}. 
Quasi Hyperbolic methods are capable of recovering Accelerated SGD \cite{jain2017acc}, Nesterov Accelerated Gradient \cite{nesterov1983nag}, Synthesized Nesterov Variants \cite{less2016snv}, and others, thus covering more algorithms than those present. However, these methods are included primarily for reference, and the major focus of this work is on the schedules describe at the beginning of Section \ref{experiments}.
\emph{We emphasize that \textsc{Demon} can be combined with any momentum method. We present results with SGDM and Adam due to their wide usage.}


\medskip
\noindent \textbf{Mainline results} We summarize the results of all relevant settings in Table \ref{tablepctresults}.
Out of the 28 relevant settings in this paper, \textsc{Demon} achieves the highest percentage of Top-1 and Top-3 finishes, with 39\% and 85\% respectively.
Other momentum decay schedules are not competitive, likely due to being overly aggressive.
Learning rate step schedule, linear schedule, and cosine schedule perform comparably; however, the different learning rate schedulers do not always yield comparable performance across settings.
For example, linear learning rate decay performs exceptionally well in the ResNet settings, but closer to average on other settings.
Such results indicate that the decay schedule, for learning rate or momentum, is an additional hyperparameter that must be tuned.
Even though Decay On Plateau and Exponential Decay are implemented in most popular frameworks as empirical alternatives, they perform poorly when the total number of epochs are predefined.
\textsc{Demon} is the most consistent across a wide range of settings, with by far the highest Top-3 performance.

\medskip
\noindent \textbf{Residual Networks} (\texttt{RN20-CIFAR10}). We train a ResNet20 \cite{he2016deep} model on the CIFAR-10 dataset.
We emphasize that ResNet20 is commonly conflated with the more expressive ResNet18, which achieves different performance.
This is an important setting to evaluate because ResNets remain one of the most popular computer vision architectures in both academia and industry, achieving reasonable performance with less risk of overfitting to the test set \cite{pmlr-v119-shankar20c}.
The other ResNet settings include \texttt{WRN-STL10}, a Wide Residual 16-8 model \cite{zagoruyko2016wide} on STL-10 that has significantly fewer, higher resolution images in comparison to CIFAR, and \texttt{RN56-TINYIMAGENET}, a ResNet56 model on the Tiny ImageNet dataset. Due to limited resources, the runs of \texttt{RN56-TINYIMAGENET} are conducted to supplement the other settings.
See Table \ref{tableresnets} for results.

In sum, the other momentum decay schedules do not perform well, likely due to the overly aggressive momentum decay.
The momentum component of OneCycle, when used in isolation, appears to destabilize training, leading in some cases to performance worse than the vanilla counterpart.
Whilst traditionally the learning rate step schedule is the most popular and often used to achieve state-of-the-art results in computer vision, this schedule has no clear advantage over the learning rate cosine schedule, the learning rate linear schedule, or \textsc{Demon}.
\textsc{Demon} has the strongest performance, even outperforming methods that automatically tune the momentum parameter.

\begin{figure*}
  \centering
    \begin{subfigure}[b]{0.33\linewidth}
    \includegraphics[width=\linewidth]{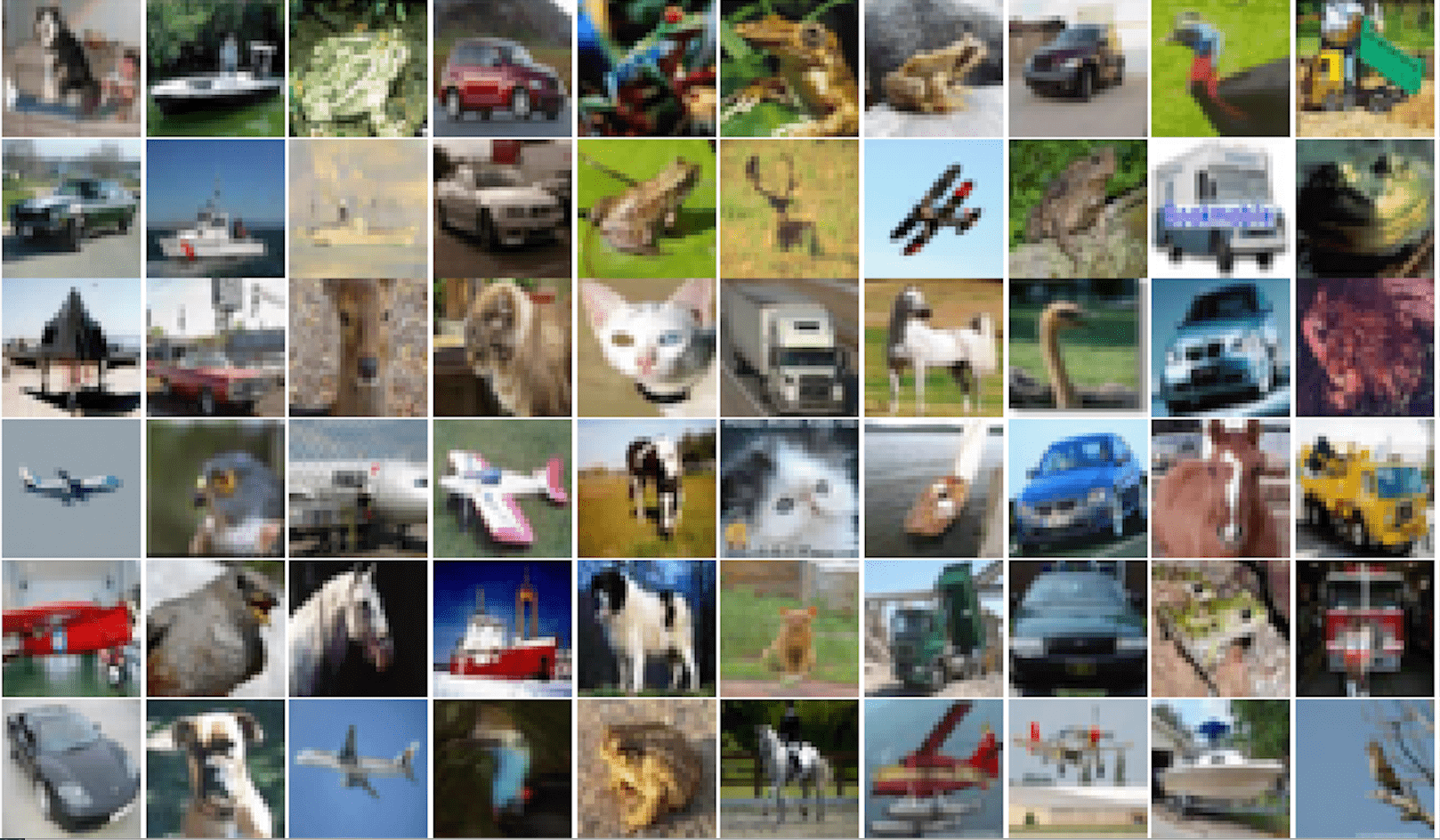}
    \label{fig:ncsnCifar10Sample}
  \end{subfigure}
  \begin{subfigure}[b]{0.32\linewidth}
    \includegraphics[width=\linewidth]{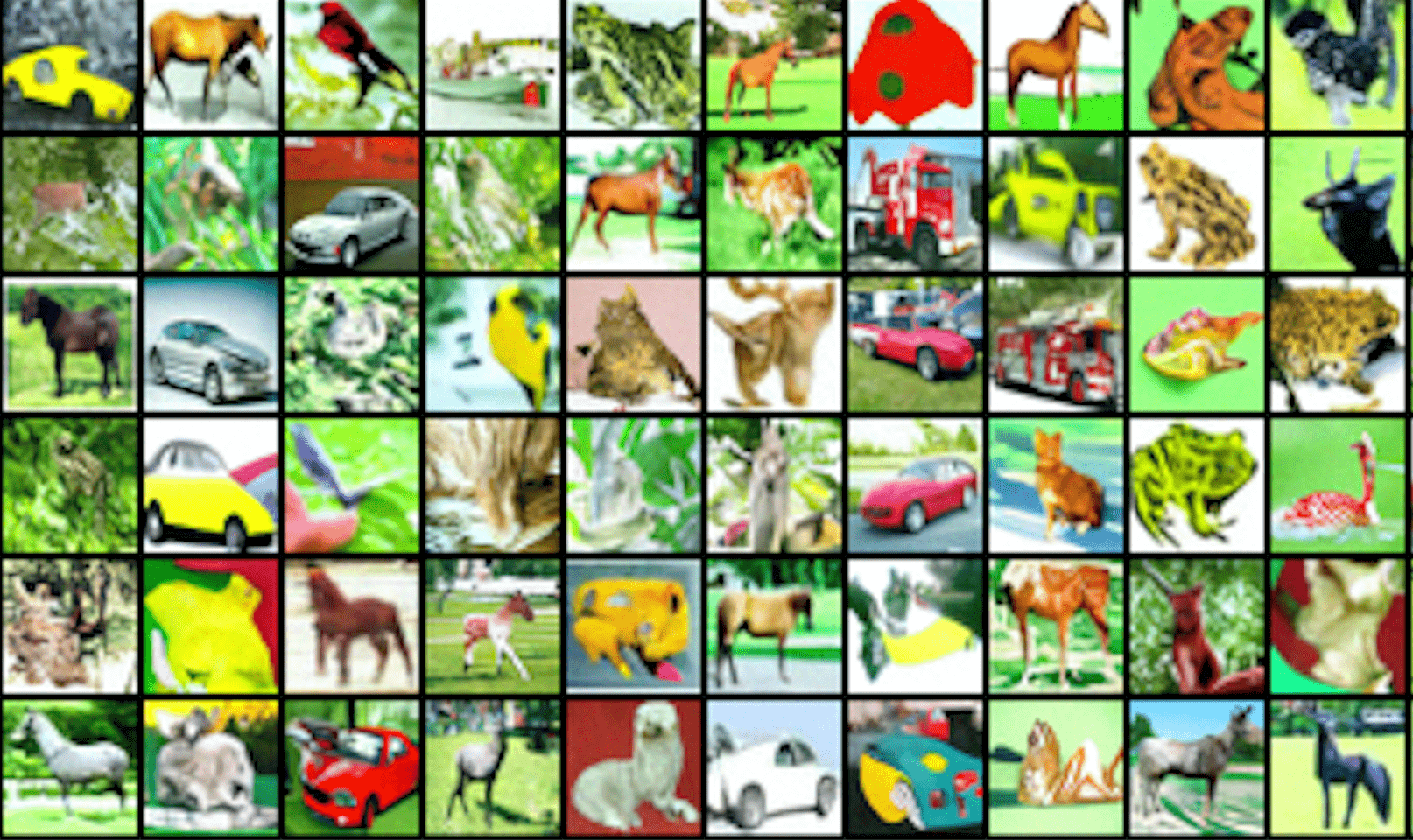}
    \label{fig:ncsnAdamSample}
  \end{subfigure} 
  \begin{subfigure}[b]{0.32\linewidth}
    \includegraphics[width=\linewidth]{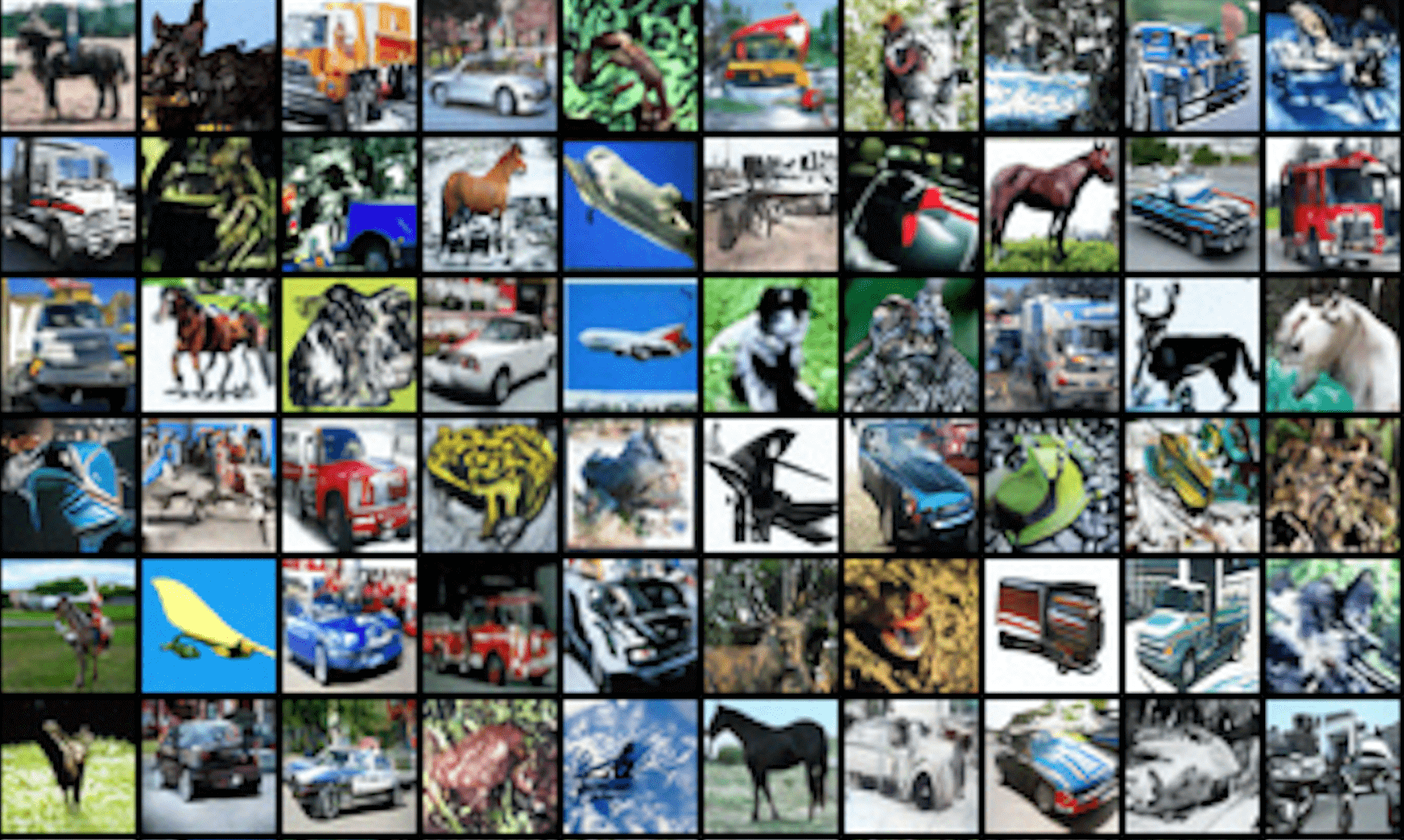}
    \label{fig:ncsnDemonAdamSample}
  \end{subfigure}
  \caption{\small Randomly selected CIFAR10 images generated with NCSN. Left: Real CIFAR10 images. Middle: Adam. Right: \textsc{Demon} Adam.}
  \label{fig:ncsnResults}
\end{figure*}

\begin{table*}[t]
\centering
\begin{small}
\caption{\texttt{VGG16-CIFAR100} generalization error, \texttt{LSTM-PTB} generalization perplexity, \texttt{VAE-MNIST} generalization loss, and \texttt{CAPS-FMNIST} generalization error. The number of epochs was predefined before the execution of the algorithms. Red indicates Top-1 performance, bold is Top-3, ignoring non SGDM and Adam optimizers.}
\begin{tabular}{c|cc|cc|cc|cc}
\toprule
    & \multicolumn{2}{c}{\textbf{VGG-16}} & \multicolumn{2}{c}{\textbf{LSTM}} & \multicolumn{2}{c}{\textbf{VAE}} & \multicolumn{2}{c}{\textbf{CAPSNET}} \\ \midrule
    SGDM & 150 epochs  & 300 epochs & 25 epochs & 39 epochs & 50 epochs & 100 epochs & 50 epochs & 100 epochs \\ \midrule
    + LR Step Schedule & 30.09 $\pm$ .32 & \textbf{27.83} $\pm$ .30 & \textbf{81.67} $\pm$ .21 & \textbf{82.02} $\pm$ .13 & 140.28 $\pm$ .51 & 137.70 $\pm$ .93 & - & - \\
    + LR Cosine Schedule  & \textcolor{myred}{\textbf{28.63}} $\pm$ .11 & \textbf{27.84} $\pm$ .12 & \textbf{81.64} $\pm$ .37 & \textbf{83.23} $\pm$ .06 & \textbf{139.15} $\pm$ .26  & \textbf{136.69} $\pm$ .27 & - & - \\    
    + OneCycle  & 30.10 $\pm$ .34 & 29.09 $\pm$ .12 & 90.03 $\pm$ .39 & 91.19 $\pm$ .01 & \textbf{139.79} $\pm$ .66 & \textbf{137.20} $\pm$ .06 & - & - \\  
    + LR Linear Schedule  & \textbf{29.10} $\pm$ .34 & 28.26 $\pm$ .08 & 96.27 $\pm$ .09 &  98.79 $\pm$ .02 & 148.00 $\pm$ .48 & 141.72 $\pm$ .48 &  - & -   \\ 
    + LR Decay on Plateau  & 30.65 $\pm$ .31 & 29.74 $\pm$ .43 & \textcolor{myred}{\textbf{81.55}} $\pm$ .24 & \textcolor{myred}{\textbf{81.82}} $\pm$ .07 & 140.51 $\pm$ .73 & 139.54 $\pm$ .34 & - & -  \\ 
    + LR Exp decay  & 29.51 $\pm$ .22  & 28.47 $\pm$ .18 & 84.20 $\pm$ .08 & 83.49 $\pm$ .03 & 154.31 $\pm$ .43 & 145.83 $\pm$ .48 & - & - \\ 
    + OneCycle Momentum  & 35.86 $\pm$ .25 & 35.34 $\pm$ .30 & 87.14 $\pm$ .27 & 91.93 $\pm$ 1.03 & 144.90 $\pm$ .61 & 142.63 $\pm$ .25 & - & -  \\ 
    + Cosine Momentum  & 32.73 $\pm$ .07 & 30.99 $\pm$ .11 & 88.33 $\pm$ .92 & 90.02 $\pm$ .10 & 145.13 $\pm$ .85 & 140.14 $\pm$ .81 & - & - \\ 
    + Linear Momentum  & 31.61 $\pm$ .29 & 31.23 $\pm$ .26 & 90.02 $\pm$ .51 & 93.06 $\pm$ .29 & 145.33 $\pm$ .13 & 139.83 $\pm$ .14 & - & - \\ 
    + Exp Momentum decay  & 34.50 $\pm$ .23 & 32.83 $\pm$ .13 & 86.39 $\pm$ .20 & 89.45 $\pm$ .63 & 150.54 $\pm$ 1.07 & 156.95 $\pm$ .47 & - & -   \\ \midrule
    AggMo + LR Step & 30.75 $\pm$ .55 & 28.64 $\pm$ .45 & 83.15 $\pm$ .12 & 83.43 $\pm$ .17 & 139.49 $\pm$ .99 & 136.56 $\pm$ .28 & - & - \\
    QHM + LR Step & 29.93 $\pm$ .13 & 29.01 $\pm$ .54 & 88.75 $\pm$ .23 & 88.42 $\pm$ .10 & 142.47 $\pm$ .50 & 137.97 $\pm$ .54 & - & - \\ \midrule
    \textsc{Demon} SGDM & \textbf{28.67} $\pm$ .11 & \textcolor{myred}{\textbf{27.69}} $\pm$ .11 & 82.66 $\pm$ .05 & 84.84 $\pm$ .22 & \textcolor{myred}{\textbf{138.29}} $\pm$ .08 & \textcolor{myred}{\textbf{136.55}} $\pm$ .64 & - & - \\ \midrule \midrule
    Adam & 33.62 $\pm$ .11 & 31.09 $\pm$ .09 & 109.48 $\pm$ .36 & 116.26 $\pm$ .10 & 136.28 $\pm$ .18 & 134.64 $\pm$ .14 & 9.27 $\pm$ .08  & 9.25 $\pm$ .11\\
    + LR Step Schedule & 29.40 $\pm$ .22 & \textbf{27.75} $\pm$ .15 & \textbf{98.79} $\pm$ .05 & \textcolor{myred}{\textbf{99.69}} $\pm$ .03 & 136.62 $\pm$ .30 & 134.14 $\pm$ .56 & \textbf{8.90} $\pm$ .03 & 8.98 $\pm$ .05 \\  
    + LR Cosine Schedule  & 29.68 $\pm$ .17 & \textbf{28.08}  $\pm$ .10 & \textcolor{myred}{\textbf{97.70}} $\pm$ .07  & \textbf{100.56} $\pm$ .27 & 134.73 $\pm$ .04 & \textcolor{myred}{\textbf{133.25}} $\pm$ .26 &  9.16 $\pm$ .07 & 9.75 $\pm$ .10 \\    
    + OneCycle  & 29.83 $\pm$ .29 & 29.58 $\pm$ .18 & 112.32 $\pm$ .10 &  117.31 $\pm$ .11 & \textbf{134.67} $\pm$ .55 & \textbf{133.27} $\pm$ .07 & 9.21 $\pm$ .04  & 8.88 $\pm$ .04 \\  
    + LR Linear Schedule  & \textbf{29.30} $\pm$ .18 & 28.65 $\pm$ .10 & 111.10 $\pm$ .83  & 115.24 $\pm$ .10 & \textbf{134.71} $\pm$ .25 & 134.00 $\pm$ .49 & \textbf{8.90} $\pm$ .10 &  8.90 $\pm$ .04 \\ 
    + LR Decay on Plateau  & \textbf{29.03} $\pm$ .10 & 28.67 $\pm$ .19 & \textbf{99.65} $\pm$ .18 &  \textbf{101.46} $\pm$ .22 & 135.68 $\pm$ .59 & 134.10 $\pm$ .21  & 9.10 $\pm$ .09 & 9.12 $\pm$ .07 \\ 
    + LR Exp decay  & 29.53 $\pm$ .12 & 28.83 $\pm$ .08 & 103.50 $\pm$ .41 & 103.09 $\pm$ .10 & 135.19 $\pm$ .43 & 134.05 $\pm$ .16 & \textcolor{myred}{\textbf{8.80}} $\pm$ .07 &  \textcolor{myred}{\textbf{8.72}} $\pm$ .08 \\ 
    + OneCycle Momentum  & 36.98 $\pm$ .29 & 32.96 $\pm$ .37 & 106.86 $\pm$ .66  & 108.77 $\pm$ .08 & 138.16 $\pm$ .35 & 136.68 $\pm$ .47 & 9.79 $\pm$ .09 &  9.95 $\pm$ .08 \\ 
    + Cosine Momentum  & 31.65 $\pm$ .16 & 30.54 $\pm$ .16 & 105.78 $\pm$ .04 & 106.01 $\pm$ .09 & 135.62 $\pm$ .38 & 134.02 $\pm$ .10 & 9.13 $\pm$ .06 & \textbf{8.82} $\pm$ .10\\ 
    + Linear Momentum  & 32.28 $\pm$ .13 & 29.65 $\pm$ .11 & 106.38 $\pm$ .49 & 114.24 $\pm$ .14 & 136.24 $\pm$ .87 & 135.00 $\pm$ .51 & 9.12 $\pm$ .08 & 9.03 $\pm$ .04 \\  
    + Exp Momentum decay  & 32.05 $\pm$ .14 &  30.63 $\pm$ .14 & 104.32 $\pm$ .26 & 115.87 $\pm$ .12 & 136.67 $\pm$ .56 & 136.05 $\pm$ .11 & 9.00 $\pm$ .07  & 9.22 $\pm$ .08  \\ \midrule
    AMSGrad & 34.46 $\pm$ .21 & 31.62 $\pm$ .12 & 103.12 $\pm$ .18 & 103.26 $\pm$ .17 & 137.89 $\pm$ .12 & 135.69 $\pm$ .03 & 9.39 $\pm$ .18 & 9.28 $\pm$ .19\\
    AdamW & 33.48 $\pm$ .68 & 32.22 $\pm$ .13 & 110.10 $\pm$ 1.32 & 110.72 $\pm$ 1.63 & 136.15 $\pm$ .21 & 134.68 $\pm$ .19 & 9.78 $\pm$ .62 & 9.92 $\pm$ .74 \\
    QHAdam & 32.96 $\pm$ .11 & 30.97 $\pm$ .10 & 106.98 $\pm$ .25 & 106.18 $\pm$ .32 & 136.69 $\pm$ .17 & 134.84 $\pm$ .08 & 9.30 $\pm$ .23 & 9.24 $\pm$ .15\\
    YellowFin & 68.87 $\pm$ 5.82 & 50.18 $\pm$ 4.02 & 117.21 $\pm$ .42 & 109.04 $\pm$ .20 & 414.74 $\pm$ 5.00 & 351.80 $\pm$ 6.68 & 10.96 $\pm$ .65 & 10.55 $\pm$ .84 \\ \midrule
    \textsc{Demon} Adam & \textcolor{myred}{\textbf{28.84}} $\pm$ .18 & \textcolor{myred}{\textbf{27.11}} $\pm$ .19 & 104.60 $\pm$ .16 &  107.07 $\pm$ .05 & \textcolor{myred}{\textbf{134.35}} $\pm$ .24 & \textbf{133.98} $\pm$ .40 & \textbf{8.88} $\pm$ .05 & \textbf{8.87} $\pm$ .10 \\
 \bottomrule
\end{tabular}
\label{tablevgglstmvae}
\end{small}
\end{table*}

\medskip
\noindent \textbf{Non-Residual Convolutional Network} (\texttt{VGG16-CIFAR100}). 
We train an adjusted VGG-16 model \cite{simonyan2014very} on the CIFAR-100 dataset. Previous observations from the ResNet settings continue to hold, where other momentum decay methods are not competitive. Again, the learning rate step schedule does not appear to provide any advantage over the cosine schedule, linear schedule, or \textsc{Demon}, the latter of which performs the best. See Table \ref{tablevgglstmvae}.

\medskip
\noindent \textbf{LSTM} \texttt{(PTB-LSTM)}. 
We apply an LSTM \cite{hochreiter1997long} architecture to the language modeling task, which has notoriously sharp gradient distributions (e.g., rare words).
We use the official TensorFlow v1 implementation for PTB - LSTM.
OneCycle Momentum is adjusted to decay to 0 and back since this setting typically requires low momentum to train well.
Such characteristically low momentum makes this task a difficult test case for momentum decay methods, as momentum seems to have less impact on performance relative to other settings. 
However, more momentum values are also swept to achieve the reported perplexity.
As a whole, smooth decay methods do not perform comparably to step decay methods on this task.
\textsc{Demon} is surprisingly competitive and is the only momentum method to achieve reasonable performance with both SGDM and Adam. See Table \ref{tablevgglstmvae}.

\medskip
\noindent \textbf{Variational AutoEncoder}\texttt{(VAE-MNIST)}. Generative modeling is a branch of unsupervised learning that focuses on learning the underlying data distribution. VAEs \cite{kingma2015vae} are generative models that pair a generator network with a recognition model that performs approximate inference and can be trained with backprop. We train VAEs on MNIST. General trends follow the ResNet settings. Interestingly, the learning rate linear schedule perform poorly with SGDM, but are improved in the Adam setting. See Table \ref{tablevgglstmvae}.



\medskip
\noindent \textbf{Capsule Network} \texttt{(CAPS-FMNIST)}. Capsule Networks \cite{sabour2017caps} represent Neural Networks as a set of capsules that each encode a specific entity or meaning. Capsules exploit the observation that viewpoint changes significantly alter pixels but are linear with respect to the pose matrix. The activation of capsules differs from standard neural network activation functions because it depends on comparing incoming pose predictions. We train Capsule Networks on the FMNIST dataset with only Adam and its variants, which are typically used in this setting\cite{sabour2017caps}. Highlighting the unpredictability of the performance among learning rate schedules, the exponential learning rate decay schedule is unremarkable in other settings, but is clearly the best learning rate schedule in this setting. See Table \ref{tablevgglstmvae}.


\medskip
\noindent \textbf{Noise Conditional Score Network}\texttt{(NCSN-CIFAR10}. NCSN \cite{song2019generative} is a recent generative model that estimates gradients of the data distribution with score matching and produces samples via Langevin dynamics. We train a NCSN on CIFAR10, for which NCSN achieves a strong inception score. Although vanilla Adam achieves a slightly superior inception score (see Table \ref{tablencsn}) the results in Figure \ref{fig:ncsnResults} are unnaturally green compared to those produced by \textsc{Demon} Adam.

\begin{table}
\centering
\caption{\small \texttt{NCSN-CIFAR10} inception score. The number of epochs was predefined before the execution of the algorithms.}
\begin{tabular}{c|c} \toprule
    &  {\textbf{NCSN}} (512 epochs) \\ \midrule
    Adam & \textcolor{myred}{\textbf{8.15}} $\pm$ .20 \\
    \textsc{Demon} Adam & 8.07 $\pm$ .08\\ 
    \bottomrule
\end{tabular}
\label{tablencsn}
\end{table}

\begin{table*}
\centering
\caption{\small Results of \texttt{BERT$_{BASE}$-GLUE}. Adam + LR Linear Schedule follows the huggingface \cite{wolf2020huggingfaces} implementation, and achieves the results in well-known studies \cite{devlin2019bert, sanh2020distilbert}. }
\begin{tabular}{c|c|ccccccccc} 
\toprule
    & Score & CoLA & MNLI & MRPC & QNLI  & QQP & RTE & SST-2 & STS-B & WNLI \\ \midrule
    Adam + LR Linear Schedule & 79.1 & 57.4 & 84.3 & 89.0 & 91.4 & 89.2 & 69.4 & 92.7 & 89.0 & 49.6\\
    \textsc{Demon} Adam & \textbf{79.7} & 58.4 & 84.2 & 90.0 & 90.9 & 89.0 & 69.4 & 92.5 & 88.8 & 53.8\\
 \bottomrule
\end{tabular}
\label{tablebertgluecomp}
\end{table*}

\begin{table*}
\centering
\begin{footnotesize}
\caption{\small \texttt{VGG16-CIFAR100-DEMONSGDM}, \texttt{RN20-CIFAR10-DEMONSGDM} generalization error, \texttt{PTB-LSTM-DEMONSGDM} (perplexity) and \texttt{VAE-MNIST-DEMONSGDM} generalization loss. The number of epochs was predefined before the execution. }
\begin{tabular}{c|ccc|ccc|cc|cc} 
\toprule
    & \multicolumn{3}{c}{\textbf{RN-20}} & \multicolumn{3}{c}{\textbf{VGG-16}} & \multicolumn{2}{c}{\textbf{LSTM}} & \multicolumn{2}{c}{\textbf{VAE}}\\ \midrule
    & 75 epochs & 150 epochs  & 300 epochs & 50 epochs & 100 epochs  & 200 epochs & 25 epochs & 39 epochs & 50 epochs & 100 epochs\\ \midrule
    SGD ELR &  9.14 $\pm$ .24 & 8.58 $\pm$ .08 & 8.16 $\pm$ .15 & 35.71 $\pm$ .54 & 30.11 $\pm$ .28 & 29.21 $\pm$ .32 & inf & inf & inf & inf\\
    \textsc{Demon} SGDM & \textcolor{myred}{\textbf{8.71}} $\pm$ .24 & \textcolor{myred}{\textbf{7.95}} $\pm$ .15 & \textcolor{myred}{\textbf{7.59}} $\pm$ .12 & \textcolor{myred}{\textbf{32.26}} $\pm$ .21 & \textcolor{myred}{\textbf{28.67}} $\pm$ .11 & \textcolor{myred}{\textbf{27.69}} $\pm$ .11 &\textcolor{myred}{\textbf{82.66}} $\pm$ .105& \textcolor{myred}{\textbf{84.84}} $\pm$ .22 & \textcolor{myred}{\textbf{138.29}} $\pm$ .08 & \textcolor{myred}{\textbf{136.55}} $\pm$ .64 \\
 \bottomrule
\end{tabular}
\label{tablevgg16wrnelrsgdcomp}
\end{footnotesize}
\end{table*}

\medskip
\noindent \textbf{BERT}\texttt{(BERT$_{BASE}$-GLUE)}. BERT \cite{devlin2019bert} is one of the most influential language models in the last few years. The key characteristic of BERT is the ability for a pre-trained model to be fine-tuned to achieve strong performance across a large variety of language tasks. The GLUE benchmark \cite{wang2019glue} is a collection of nine different language tasks \cite{warstadt2018neural, socher2013recursive, dolan2005automatically, agirre2007semantic, williams2018broad, rajpurkar2016squad, dagan2006pascal, bar2006second, giampiccolo2007third, bentivogli2009fifth, levesque2011winograd}, and is a common benchmark in NLP. We achieve the performance in well-known studies \cite{devlin2019bert, sanh2020distilbert}, using the huggingface \cite{wolf2020huggingfaces} default implementation of Adam with learning rate linear decay, tuning only the learning rate. 
We also only tune the learning rate for \textsc{Demon} Adam. Results are given in Table \ref{tablebertgluecomp}. Running the same seeds, \textsc{Demon} Adam yields a slight improvement over the baseline.

\section{Demon and Effective Learning Rate}\label{effectiveLearningRateAppendix}
We present results of Demon against the effective learning rate adjusted SGD (SGD ELR). The effective learning rate is proposed to approximate SGDM with SGD, where the learning rate is adjusted with a factor of $1/(1 - m)$ and $m$ is the momentum coefficient. However, the results in Tables \ref{tablevgg16wrnelrsgdcomp} demonstrate that $\textsc{Demon}$ cannot be accurately approximated with an effective learning rate adjusted SGD. For settings (\texttt{PTB-LSTM-DEMONSGDM} and \texttt{VAE-MNIST-DEMONSGDM}), SGD ELR causes learning to diverge. In Table \ref{tablevgg16wrnelrsgdcomp}, there exists a 0.5-3\% generalization error gap for \texttt{VGG16-CIFAR100} and for \texttt{RN20-CIFAR10}.

\section{Conclusion}

We show the effectiveness of \textsc{Demon} across a large number of datasets and architectures. We demonstrate that \textsc{Demon} can be effectively applied to both SGDM and Adam. Compared to other learning rate schedules and momentum schedules, \textsc{Demon} achieves the largest number of Top-1 and Top-3 finishes. This includes improvements over the popular learning rate step schedule, cosine decay schedule, OneCycle, and many others. \textsc{Demon} is computationally cheap, understandable, and easy to implement. We hope it is useful in practice and as a subject of future research

\bibliographystyle{ACM-Reference-Format}
\bibliography{iclr2021_conference}


\begin{thebibliography}{82}


\ifx \showCODEN    \undefined \def \showCODEN     #1{\unskip}     \fi
\ifx \showDOI      \undefined \def \showDOI       #1{#1}\fi
\ifx \showISBNx    \undefined \def \showISBNx     #1{\unskip}     \fi
\ifx \showISBNxiii \undefined \def \showISBNxiii  #1{\unskip}     \fi
\ifx \showISSN     \undefined \def \showISSN      #1{\unskip}     \fi
\ifx \showLCCN     \undefined \def \showLCCN      #1{\unskip}     \fi
\ifx \shownote     \undefined \def \shownote      #1{#1}          \fi
\ifx \showarticletitle \undefined \def \showarticletitle #1{#1}   \fi
\ifx \showURL      \undefined \def \showURL       {\relax}        \fi
\providecommand\bibfield[2]{#2}
\providecommand\bibinfo[2]{#2}
\providecommand\natexlab[1]{#1}
\providecommand\showeprint[2][]{arXiv:#2}

\bibitem[\protect\citeauthoryear{Abadi, Barham, Chen, Chen, Davis, Dean, Devin,
  Ghemawat, Irving, Isard, Kudlur, Levenberg, Monga, Moore, Murray, Steiner,
  Tucker, Vasudevan, Warden, Wicke, Yu, and Zheng}{Abadi et~al\mbox{.}}{2016}]%
        {45381}
\bibfield{author}{\bibinfo{person}{Martin Abadi}, \bibinfo{person}{Paul
  Barham}, \bibinfo{person}{Jianmin Chen}, \bibinfo{person}{Zhifeng Chen},
  \bibinfo{person}{Andy Davis}, \bibinfo{person}{Jeffrey Dean},
  \bibinfo{person}{Matthieu Devin}, \bibinfo{person}{Sanjay Ghemawat},
  \bibinfo{person}{Geoffrey Irving}, \bibinfo{person}{Michael Isard},
  \bibinfo{person}{Manjunath Kudlur}, \bibinfo{person}{Josh Levenberg},
  \bibinfo{person}{Rajat Monga}, \bibinfo{person}{Sherry Moore},
  \bibinfo{person}{Derek~G. Murray}, \bibinfo{person}{Benoit Steiner},
  \bibinfo{person}{Paul Tucker}, \bibinfo{person}{Vijay Vasudevan},
  \bibinfo{person}{Pete Warden}, \bibinfo{person}{Martin Wicke},
  \bibinfo{person}{Yuan Yu}, {and} \bibinfo{person}{Xiaoqiang Zheng}.}
  \bibinfo{year}{2016}\natexlab{}.
\newblock \showarticletitle{TensorFlow: A system for large-scale machine
  learning}. In \bibinfo{booktitle}{\emph{12th USENIX Symposium on Operating
  Systems Design and Implementation (OSDI 16)}}. \bibinfo{pages}{265--283}.
\newblock
\urldef\tempurl%
\url{https://www.usenix.org/system/files/conference/osdi16/osdi16-abadi.pdf}
\showURL{%
\tempurl}


\bibitem[\protect\citeauthoryear{Agarwal, Anil, Hazan, Koren, and
  Zhang}{Agarwal et~al\mbox{.}}{2020}]%
        {revis_opt}
\bibfield{author}{\bibinfo{person}{Naman Agarwal}, \bibinfo{person}{Rohan
  Anil}, \bibinfo{person}{Elad Hazan}, \bibinfo{person}{Tomer Koren}, {and}
  \bibinfo{person}{Cyril Zhang}.} \bibinfo{year}{2020}\natexlab{}.
\newblock \showarticletitle{Revisiting the Generalization of Adaptive Gradient
  Methods}.
\newblock  (\bibinfo{year}{2020}).
\newblock


\bibitem[\protect\citeauthoryear{Agirre, M`arquez, and Wicentowski}{Agirre
  et~al\mbox{.}}{2007}]%
        {agirre2007semantic}
\bibfield{editor}{\bibinfo{person}{Eneko Agirre}, \bibinfo{person}{Llu'{i}s
  M`arquez}, {and} \bibinfo{person}{Richard Wicentowski}} (Eds.).
  \bibinfo{year}{2007}\natexlab{}.
\newblock \bibinfo{booktitle}{\emph{Proceedings of the Fourth International
  Workshop on Semantic Evaluations (SemEval-2007)}}.
\newblock \bibinfo{publisher}{Association for Computational Linguistics},
  \bibinfo{address}{Prague, Czech Republic}.
\newblock


\bibitem[\protect\citeauthoryear{Arjovsky, Chintala, and Bottou}{Arjovsky
  et~al\mbox{.}}{2017}]%
        {arjovsky2017wasserstein}
\bibfield{author}{\bibinfo{person}{Martin Arjovsky}, \bibinfo{person}{Soumith
  Chintala}, {and} \bibinfo{person}{Léon Bottou}.}
  \bibinfo{year}{2017}\natexlab{}.
\newblock \bibinfo{title}{Wasserstein GAN}.
\newblock
\newblock
\showeprint[arxiv]{1701.07875}~[stat.ML]


\bibitem[\protect\citeauthoryear{Arnold, Manzagol, Bebanezhad, Mitliagkas, and
  Roux}{Arnold et~al\mbox{.}}{2019}]%
        {arnold2019reducing}
\bibfield{author}{\bibinfo{person}{Sebastien Arnold},
  \bibinfo{person}{Pierre-Antoine Manzagol}, \bibinfo{person}{Reza Bebanezhad},
  \bibinfo{person}{Ioannis Mitliagkas}, {and} \bibinfo{person}{Nicolas~Le
  Roux}.} \bibinfo{year}{2019}\natexlab{}.
\newblock \showarticletitle{Reducing the variance in online optimization by
  transporting past gradients}. In \bibinfo{booktitle}{\emph{Advances in Neural
  Information Processing Systems}}. \bibinfo{pages}{5392--5403}.
\newblock


\bibitem[\protect\citeauthoryear{Bahdanau, Cho, and Bengio}{Bahdanau
  et~al\mbox{.}}{2014}]%
        {bahdanau2014neural}
\bibfield{author}{\bibinfo{person}{Dzmitry Bahdanau},
  \bibinfo{person}{Kyunghyun Cho}, {and} \bibinfo{person}{Yoshua Bengio}.}
  \bibinfo{year}{2014}\natexlab{}.
\newblock \showarticletitle{Neural machine translation by jointly learning to
  align and translate}.
\newblock \bibinfo{journal}{\emph{arXiv preprint arXiv:1409.0473}}
  (\bibinfo{year}{2014}).
\newblock


\bibitem[\protect\citeauthoryear{Bar~Haim, Dagan, Dolan, Ferro, Giampiccolo,
  Magnini, and Szpektor}{Bar~Haim et~al\mbox{.}}{2006}]%
        {bar2006second}
\bibfield{author}{\bibinfo{person}{Roy Bar~Haim}, \bibinfo{person}{Ido Dagan},
  \bibinfo{person}{Bill Dolan}, \bibinfo{person}{Lisa Ferro},
  \bibinfo{person}{Danilo Giampiccolo}, \bibinfo{person}{Bernardo Magnini},
  {and} \bibinfo{person}{Idan Szpektor}.} \bibinfo{year}{2006}\natexlab{}.
\newblock \showarticletitle{The second {PASCAL} recognising textual entailment
  challenge}.
\newblock  (\bibinfo{year}{2006}).
\newblock


\bibitem[\protect\citeauthoryear{Bentivogli, Dagan, Dang, Giampiccolo, and
  Magnini}{Bentivogli et~al\mbox{.}}{2009}]%
        {bentivogli2009fifth}
\bibfield{author}{\bibinfo{person}{Luisa Bentivogli}, \bibinfo{person}{Ido
  Dagan}, \bibinfo{person}{Hoa~Trang Dang}, \bibinfo{person}{Danilo
  Giampiccolo}, {and} \bibinfo{person}{Bernardo Magnini}.}
  \bibinfo{year}{2009}\natexlab{}.
\newblock \showarticletitle{The Fifth {PASCAL} Recognizing Textual Entailment
  Challenge}.
\newblock  (\bibinfo{year}{2009}).
\newblock


\bibitem[\protect\citeauthoryear{Brown, Mann, Ryder, Subbiah, Kaplan, Dhariwal,
  Neelakantan, Shyam, Sastry, Askell, Agarwal, Herbert-Voss, Krueger, Henighan,
  Child, Ramesh, Ziegler, Wu, Winter, Hesse, Chen, Sigler, Litwin, Gray, Chess,
  Clark, Berner, McCandlish, Radford, Sutskever, and Amodei}{Brown
  et~al\mbox{.}}{2020}]%
        {brown2020language}
\bibfield{author}{\bibinfo{person}{Tom~B. Brown}, \bibinfo{person}{Benjamin
  Mann}, \bibinfo{person}{Nick Ryder}, \bibinfo{person}{Melanie Subbiah},
  \bibinfo{person}{Jared Kaplan}, \bibinfo{person}{Prafulla Dhariwal},
  \bibinfo{person}{Arvind Neelakantan}, \bibinfo{person}{Pranav Shyam},
  \bibinfo{person}{Girish Sastry}, \bibinfo{person}{Amanda Askell},
  \bibinfo{person}{Sandhini Agarwal}, \bibinfo{person}{Ariel Herbert-Voss},
  \bibinfo{person}{Gretchen Krueger}, \bibinfo{person}{Tom Henighan},
  \bibinfo{person}{Rewon Child}, \bibinfo{person}{Aditya Ramesh},
  \bibinfo{person}{Daniel~M. Ziegler}, \bibinfo{person}{Jeffrey Wu},
  \bibinfo{person}{Clemens Winter}, \bibinfo{person}{Christopher Hesse},
  \bibinfo{person}{Mark Chen}, \bibinfo{person}{Eric Sigler},
  \bibinfo{person}{Mateusz Litwin}, \bibinfo{person}{Scott Gray},
  \bibinfo{person}{Benjamin Chess}, \bibinfo{person}{Jack Clark},
  \bibinfo{person}{Christopher Berner}, \bibinfo{person}{Sam McCandlish},
  \bibinfo{person}{Alec Radford}, \bibinfo{person}{Ilya Sutskever}, {and}
  \bibinfo{person}{Dario Amodei}.} \bibinfo{year}{2020}\natexlab{}.
\newblock \bibinfo{title}{Language Models are Few-Shot Learners}.
\newblock
\newblock
\showeprint[arxiv]{2005.14165}~[cs.CL]


\bibitem[\protect\citeauthoryear{Chen and Gu}{Chen and Gu}{2018}]%
        {chen2018closing}
\bibfield{author}{\bibinfo{person}{Jinghui Chen} {and}
  \bibinfo{person}{Quanquan Gu}.} \bibinfo{year}{2018}\natexlab{}.
\newblock \showarticletitle{Closing the generalization gap of adaptive gradient
  methods in training deep neural networks}.
\newblock \bibinfo{journal}{\emph{arXiv preprint arXiv:1806.06763}}
  (\bibinfo{year}{2018}).
\newblock


\bibitem[\protect\citeauthoryear{Chen, Pan, Monga, Bengio, and Jozefowicz}{Chen
  et~al\mbox{.}}{2016}]%
        {chen2016revisiting}
\bibfield{author}{\bibinfo{person}{Jianmin Chen}, \bibinfo{person}{Xinghao
  Pan}, \bibinfo{person}{Rajat Monga}, \bibinfo{person}{Samy Bengio}, {and}
  \bibinfo{person}{Rafal Jozefowicz}.} \bibinfo{year}{2016}\natexlab{}.
\newblock \showarticletitle{Revisiting distributed synchronous {SGD}}.
\newblock \bibinfo{journal}{\emph{arXiv preprint arXiv:1604.00981}}
  (\bibinfo{year}{2016}).
\newblock


\bibitem[\protect\citeauthoryear{Choi, Shallue, Nado, Lee, Maddison, and
  Dahl}{Choi et~al\mbox{.}}{2019}]%
        {emp_opt}
\bibfield{author}{\bibinfo{person}{Dami Choi}, \bibinfo{person}{Christopher~J.
  Shallue}, \bibinfo{person}{Zachary Nado}, \bibinfo{person}{Jaehoon Lee},
  \bibinfo{person}{Chris~J. Maddison}, {and} \bibinfo{person}{G. Dahl}.}
  \bibinfo{year}{2019}\natexlab{}.
\newblock \showarticletitle{On Empirical Comparisons of Optimizers for Deep
  Learning}.
\newblock \bibinfo{journal}{\emph{ArXiv}}  \bibinfo{volume}{abs/1910.05446}
  (\bibinfo{year}{2019}).
\newblock


\bibitem[\protect\citeauthoryear{Cutkosky and Orabona}{Cutkosky and
  Orabona}{2019}]%
        {cutkosky2019momentum}
\bibfield{author}{\bibinfo{person}{Ashok Cutkosky} {and}
  \bibinfo{person}{Francesco Orabona}.} \bibinfo{year}{2019}\natexlab{}.
\newblock \showarticletitle{Momentum-Based Variance Reduction in Non-Convex
  SGD}.
\newblock \bibinfo{journal}{\emph{arXiv preprint arXiv:1905.10018}}
  (\bibinfo{year}{2019}).
\newblock


\bibitem[\protect\citeauthoryear{Dagan, Glickman, and Magnini}{Dagan
  et~al\mbox{.}}{2006}]%
        {dagan2006pascal}
\bibfield{author}{\bibinfo{person}{Ido Dagan}, \bibinfo{person}{Oren Glickman},
  {and} \bibinfo{person}{Bernardo Magnini}.} \bibinfo{year}{2006}\natexlab{}.
\newblock \showarticletitle{The {PASCAL} recognising textual entailment
  challenge}.
\newblock In \bibinfo{booktitle}{\emph{Machine learning challenges. evaluating
  predictive uncertainty, visual object classification, and recognising tectual
  entailment}}. \bibinfo{publisher}{Springer}, \bibinfo{pages}{177--190}.
\newblock


\bibitem[\protect\citeauthoryear{Defazio and Gower}{Defazio and Gower}{2020}]%
        {defazio2020factorial}
\bibfield{author}{\bibinfo{person}{Aaron Defazio} {and}
  \bibinfo{person}{Robert~M. Gower}.} \bibinfo{year}{2020}\natexlab{}.
\newblock \bibinfo{title}{Factorial Powers for Stochastic Optimization}.
\newblock
\newblock
\showeprint[arxiv]{2006.01244}~[cs.LG]


\bibitem[\protect\citeauthoryear{Devlin, Chang, Lee, and Toutanova}{Devlin
  et~al\mbox{.}}{2019}]%
        {devlin2019bert}
\bibfield{author}{\bibinfo{person}{Jacob Devlin}, \bibinfo{person}{Ming-Wei
  Chang}, \bibinfo{person}{Kenton Lee}, {and} \bibinfo{person}{Kristina
  Toutanova}.} \bibinfo{year}{2019}\natexlab{}.
\newblock \bibinfo{title}{BERT: Pre-training of Deep Bidirectional Transformers
  for Language Understanding}.
\newblock
\newblock
\showeprint[arxiv]{1810.04805}~[cs.CL]


\bibitem[\protect\citeauthoryear{Dolan and Brockett}{Dolan and
  Brockett}{2005}]%
        {dolan2005automatically}
\bibfield{author}{\bibinfo{person}{William~B Dolan} {and}
  \bibinfo{person}{Chris Brockett}.} \bibinfo{year}{2005}\natexlab{}.
\newblock \showarticletitle{Automatically constructing a corpus of sentential
  paraphrases}. In \bibinfo{booktitle}{\emph{Proceedings of the International
  Workshop on Paraphrasing}}.
\newblock


\bibitem[\protect\citeauthoryear{Duchi, Hazan, and Singer}{Duchi
  et~al\mbox{.}}{2011}]%
        {duchi2011adaptive}
\bibfield{author}{\bibinfo{person}{John Duchi}, \bibinfo{person}{Elad Hazan},
  {and} \bibinfo{person}{Yoram Singer}.} \bibinfo{year}{2011}\natexlab{}.
\newblock \showarticletitle{Adaptive subgradient methods for online learning
  and stochastic optimization}.
\newblock \bibinfo{journal}{\emph{Journal of Machine Learning Research}}
  \bibinfo{volume}{12}, \bibinfo{number}{Jul} (\bibinfo{year}{2011}),
  \bibinfo{pages}{2121--2159}.
\newblock


\bibitem[\protect\citeauthoryear{Gehring, Auli, Grangier, Yarats, and
  Dauphin}{Gehring et~al\mbox{.}}{2017}]%
        {gehring2017convolutional}
\bibfield{author}{\bibinfo{person}{Jonas Gehring}, \bibinfo{person}{Michael
  Auli}, \bibinfo{person}{David Grangier}, \bibinfo{person}{Denis Yarats},
  {and} \bibinfo{person}{Yann~N Dauphin}.} \bibinfo{year}{2017}\natexlab{}.
\newblock \showarticletitle{Convolutional sequence to sequence learning}. In
  \bibinfo{booktitle}{\emph{Proceedings of the 34th International Conference on
  Machine Learning-Volume 70}}. JMLR. org, \bibinfo{pages}{1243--1252}.
\newblock


\bibitem[\protect\citeauthoryear{Ghadimi, Feyzmahdavian, and Johansson}{Ghadimi
  et~al\mbox{.}}{2014}]%
        {ghadimi2014global}
\bibfield{author}{\bibinfo{person}{Euhanna Ghadimi},
  \bibinfo{person}{Hamid~Reza Feyzmahdavian}, {and} \bibinfo{person}{Mikael
  Johansson}.} \bibinfo{year}{2014}\natexlab{}.
\newblock \showarticletitle{Global Convergence of the HeavyBall Method for
  Convex Optimization}.
\newblock \bibinfo{journal}{\emph{arXiv preprint arXiv:1412.7457}}
  (\bibinfo{year}{2014}).
\newblock


\bibitem[\protect\citeauthoryear{Giampiccolo, Magnini, Dagan, and
  Dolan}{Giampiccolo et~al\mbox{.}}{2007}]%
        {giampiccolo2007third}
\bibfield{author}{\bibinfo{person}{Danilo Giampiccolo},
  \bibinfo{person}{Bernardo Magnini}, \bibinfo{person}{Ido Dagan}, {and}
  \bibinfo{person}{Bill Dolan}.} \bibinfo{year}{2007}\natexlab{}.
\newblock \showarticletitle{The third {PASCAL} recognizing textual entailment
  challenge}. In \bibinfo{booktitle}{\emph{Proceedings of the ACL-PASCAL
  workshop on textual entailment and paraphrasing}}. Association for
  Computational Linguistics, \bibinfo{pages}{1--9}.
\newblock


\bibitem[\protect\citeauthoryear{Gidel, Hemmat, Pezeshki, Lepriol, Huang,
  Lacoste-Julien, and Mitliagkas}{Gidel et~al\mbox{.}}{2018}]%
        {gidel2018negative}
\bibfield{author}{\bibinfo{person}{Gauthier Gidel},
  \bibinfo{person}{Reyhane~Askari Hemmat}, \bibinfo{person}{Mohammad Pezeshki},
  \bibinfo{person}{Remi Lepriol}, \bibinfo{person}{Gabriel Huang},
  \bibinfo{person}{Simon Lacoste-Julien}, {and} \bibinfo{person}{Ioannis
  Mitliagkas}.} \bibinfo{year}{2018}\natexlab{}.
\newblock \showarticletitle{Negative momentum for improved game dynamics}.
\newblock \bibinfo{journal}{\emph{arXiv preprint arXiv:1807.04740}}
  (\bibinfo{year}{2018}).
\newblock


\bibitem[\protect\citeauthoryear{He, Zhang, Ren, and Sun}{He
  et~al\mbox{.}}{2016}]%
        {he2016deep}
\bibfield{author}{\bibinfo{person}{Kaiming He}, \bibinfo{person}{Xiangyu
  Zhang}, \bibinfo{person}{Shaoqing Ren}, {and} \bibinfo{person}{Jian Sun}.}
  \bibinfo{year}{2016}\natexlab{}.
\newblock \showarticletitle{Deep residual learning for image recognition}. In
  \bibinfo{booktitle}{\emph{Proceedings of the IEEE conference on computer
  vision and pattern recognition}}. \bibinfo{pages}{770--778}.
\newblock


\bibitem[\protect\citeauthoryear{Heo, Chun, Oh, Han, Yun, Kim, Uh, and Ha}{Heo
  et~al\mbox{.}}{2020}]%
        {heo2020adamp}
\bibfield{author}{\bibinfo{person}{Byeongho Heo}, \bibinfo{person}{Sanghyuk
  Chun}, \bibinfo{person}{Seong~Joon Oh}, \bibinfo{person}{Dongyoon Han},
  \bibinfo{person}{Sangdoo Yun}, \bibinfo{person}{Gyuwan Kim},
  \bibinfo{person}{Youngjung Uh}, {and} \bibinfo{person}{Jung-Woo Ha}.}
  \bibinfo{year}{2020}\natexlab{}.
\newblock \bibinfo{title}{AdamP: Slowing Down the Slowdown for Momentum
  Optimizers on Scale-invariant Weights}.
\newblock
\newblock
\showeprint[arxiv]{2006.08217}~[cs.LG]


\bibitem[\protect\citeauthoryear{Hinton, Srivastava, and Swersky}{Hinton
  et~al\mbox{.}}{2012}]%
        {hinton2012neural}
\bibfield{author}{\bibinfo{person}{Geoffrey Hinton}, \bibinfo{person}{Nitish
  Srivastava}, {and} \bibinfo{person}{Kevin Swersky}.}
  \bibinfo{year}{2012}\natexlab{}.
\newblock \showarticletitle{Neural networks for machine learning lecture 6a
  overview of mini-batch gradient descent}.
\newblock \bibinfo{journal}{\emph{Cited on}}  \bibinfo{volume}{14}
  (\bibinfo{year}{2012}), \bibinfo{pages}{8}.
\newblock


\bibitem[\protect\citeauthoryear{Hochreiter and Schmidhuber}{Hochreiter and
  Schmidhuber}{1997}]%
        {hochreiter1997long}
\bibfield{author}{\bibinfo{person}{Sepp Hochreiter} {and}
  \bibinfo{person}{J{\"u}rgen Schmidhuber}.} \bibinfo{year}{1997}\natexlab{}.
\newblock \showarticletitle{Long short-term memory}.
\newblock \bibinfo{journal}{\emph{Neural computation}} \bibinfo{volume}{9},
  \bibinfo{number}{8} (\bibinfo{year}{1997}), \bibinfo{pages}{1735--1780}.
\newblock


\bibitem[\protect\citeauthoryear{Howard, Zhu, Chen, Kalenichenko, Wang, Weyand,
  Andreetto, and Adam}{Howard et~al\mbox{.}}{2017}]%
        {howard2017mobilenets}
\bibfield{author}{\bibinfo{person}{Andrew~G Howard}, \bibinfo{person}{Menglong
  Zhu}, \bibinfo{person}{Bo Chen}, \bibinfo{person}{Dmitry Kalenichenko},
  \bibinfo{person}{Weijun Wang}, \bibinfo{person}{Tobias Weyand},
  \bibinfo{person}{Marco Andreetto}, {and} \bibinfo{person}{Hartwig Adam}.}
  \bibinfo{year}{2017}\natexlab{}.
\newblock \showarticletitle{Mobilenets: Efficient convolutional neural networks
  for mobile vision applications}.
\newblock \bibinfo{journal}{\emph{arXiv preprint arXiv:1704.04861}}
  (\bibinfo{year}{2017}).
\newblock


\bibitem[\protect\citeauthoryear{Hu, Shen, Albanie, Sun, and Wu}{Hu
  et~al\mbox{.}}{2017}]%
        {hu2017squeeze}
\bibfield{author}{\bibinfo{person}{Jie Hu}, \bibinfo{person}{Li Shen},
  \bibinfo{person}{Samuel Albanie}, \bibinfo{person}{Gang Sun}, {and}
  \bibinfo{person}{Enhua Wu}.} \bibinfo{year}{2017}\natexlab{}.
\newblock \showarticletitle{Squeeze-and-excitation networks}.
\newblock \bibinfo{journal}{\emph{arxiv preprint arXiv:1709.01507}}
  (\bibinfo{year}{2017}).
\newblock


\bibitem[\protect\citeauthoryear{Huang, Liu, Van Der~Maaten, and
  Weinberger}{Huang et~al\mbox{.}}{2017}]%
        {huang2017densely}
\bibfield{author}{\bibinfo{person}{Gao Huang}, \bibinfo{person}{Zhuang Liu},
  \bibinfo{person}{Laurens Van Der~Maaten}, {and} \bibinfo{person}{Kilian~Q
  Weinberger}.} \bibinfo{year}{2017}\natexlab{}.
\newblock \showarticletitle{Densely connected convolutional networks}. In
  \bibinfo{booktitle}{\emph{Proceedings of the IEEE conference on computer
  vision and pattern recognition}}. \bibinfo{pages}{4700--4708}.
\newblock


\bibitem[\protect\citeauthoryear{Jain, Kakade, Kidambi, Netrapalli, and
  Sidford}{Jain et~al\mbox{.}}{2017}]%
        {jain2017acc}
\bibfield{author}{\bibinfo{person}{Prateek Jain}, \bibinfo{person}{Sham~M.
  Kakade}, \bibinfo{person}{Rahul Kidambi}, \bibinfo{person}{Praneeth
  Netrapalli}, {and} \bibinfo{person}{Aaron Sidford}.}
  \bibinfo{year}{2017}\natexlab{}.
\newblock \showarticletitle{Accelerating Stochastic Gradient Descent For Least
  Squares Regression}.
\newblock \bibinfo{journal}{\emph{arXiv preprint arXiv:1704.08227}}
  (\bibinfo{year}{2017}).
\newblock


\bibitem[\protect\citeauthoryear{Kidambi, Netrapalli, Jain, and Kakade}{Kidambi
  et~al\mbox{.}}{2018}]%
        {kidambi2018insufficiency}
\bibfield{author}{\bibinfo{person}{Rahul Kidambi}, \bibinfo{person}{Praneeth
  Netrapalli}, \bibinfo{person}{Prateek Jain}, {and} \bibinfo{person}{Sham
  Kakade}.} \bibinfo{year}{2018}\natexlab{}.
\newblock \showarticletitle{On the insufficiency of existing momentum schemes
  for stochastic optimization}. In \bibinfo{booktitle}{\emph{2018 Information
  Theory and Applications Workshop (ITA)}}. IEEE, \bibinfo{pages}{1--9}.
\newblock


\bibitem[\protect\citeauthoryear{Kingma and Ba}{Kingma and Ba}{2014}]%
        {kingma2014adam}
\bibfield{author}{\bibinfo{person}{Diederik~P Kingma} {and}
  \bibinfo{person}{Jimmy Ba}.} \bibinfo{year}{2014}\natexlab{}.
\newblock \showarticletitle{Adam: A method for stochastic optimization}.
\newblock \bibinfo{journal}{\emph{arXiv preprint arXiv:1412.6980}}
  (\bibinfo{year}{2014}).
\newblock


\bibitem[\protect\citeauthoryear{Kingma and Welling}{Kingma and
  Welling}{2015}]%
        {kingma2015vae}
\bibfield{author}{\bibinfo{person}{Diederik~P Kingma} {and}
  \bibinfo{person}{Max Welling}.} \bibinfo{year}{2015}\natexlab{}.
\newblock \showarticletitle{Auto-encoding variational Bayes.}
\newblock \bibinfo{journal}{\emph{arXiv preprint arXiv:1312.6114}}
  (\bibinfo{year}{2015}).
\newblock


\bibitem[\protect\citeauthoryear{Krizhevsky, Sutskever, and Hinton}{Krizhevsky
  et~al\mbox{.}}{2012}]%
        {krizhevsky2012imagenet}
\bibfield{author}{\bibinfo{person}{Alex Krizhevsky}, \bibinfo{person}{Ilya
  Sutskever}, {and} \bibinfo{person}{Geoffrey~E Hinton}.}
  \bibinfo{year}{2012}\natexlab{}.
\newblock \showarticletitle{Imagenet classification with deep convolutional
  neural networks}. In \bibinfo{booktitle}{\emph{Advances in neural information
  processing systems}}. \bibinfo{pages}{1097--1105}.
\newblock


\bibitem[\protect\citeauthoryear{Lessard, Recht, and Packard}{Lessard
  et~al\mbox{.}}{2016}]%
        {less2016snv}
\bibfield{author}{\bibinfo{person}{Laurent Lessard}, \bibinfo{person}{Benjamin
  Recht}, {and} \bibinfo{person}{Andrew Packard}.}
  \bibinfo{year}{2016}\natexlab{}.
\newblock \showarticletitle{Analysis and design of optimization algorithms via
  integral quadratic constraints}.
\newblock \bibinfo{journal}{\emph{SIAM Journal on Optimization}}
  \bibinfo{volume}{26}, \bibinfo{number}{1} (\bibinfo{year}{2016}),
  \bibinfo{pages}{57--95}.
\newblock


\bibitem[\protect\citeauthoryear{Levesque, Davis, and Morgenstern}{Levesque
  et~al\mbox{.}}{2011}]%
        {levesque2011winograd}
\bibfield{author}{\bibinfo{person}{Hector~J Levesque}, \bibinfo{person}{Ernest
  Davis}, {and} \bibinfo{person}{Leora Morgenstern}.}
  \bibinfo{year}{2011}\natexlab{}.
\newblock \showarticletitle{The {W}inograd schema challenge}. In
  \bibinfo{booktitle}{\emph{{AAAI} Spring Symposium: Logical Formalizations of
  Commonsense Reasoning}}, Vol.~\bibinfo{volume}{46}. \bibinfo{pages}{47}.
\newblock


\bibitem[\protect\citeauthoryear{Lin, Dollar, Girshick, He, Hariharan, and
  Belongie}{Lin et~al\mbox{.}}{2017}]%
        {lin2017feature}
\bibfield{author}{\bibinfo{person}{Tsung-Yi Lin}, \bibinfo{person}{Piotr
  Dollar}, \bibinfo{person}{Ross Girshick}, \bibinfo{person}{Kaiming He},
  \bibinfo{person}{Bharath Hariharan}, {and} \bibinfo{person}{Serge Belongie}.}
  \bibinfo{year}{2017}\natexlab{}.
\newblock \showarticletitle{Feature pyramid networks for object detection}.
\newblock \bibinfo{journal}{\emph{CVPR}} (\bibinfo{year}{2017}).
\newblock


\bibitem[\protect\citeauthoryear{Loshchilov and Hutter}{Loshchilov and
  Hutter}{2017a}]%
        {loshchilov2017fixing}
\bibfield{author}{\bibinfo{person}{Ilya Loshchilov} {and}
  \bibinfo{person}{Frank Hutter}.} \bibinfo{year}{2017}\natexlab{a}.
\newblock \showarticletitle{Fixing weight decay regularization in adam}.
\newblock \bibinfo{journal}{\emph{arXiv preprint arXiv:1711.05101}}
  (\bibinfo{year}{2017}).
\newblock


\bibitem[\protect\citeauthoryear{Loshchilov and Hutter}{Loshchilov and
  Hutter}{2017b}]%
        {loshchilov2017sgdr}
\bibfield{author}{\bibinfo{person}{Ilya Loshchilov} {and}
  \bibinfo{person}{Frank Hutter}.} \bibinfo{year}{2017}\natexlab{b}.
\newblock \bibinfo{title}{SGDR: Stochastic Gradient Descent with Warm
  Restarts}.
\newblock
\newblock
\showeprint[arxiv]{1608.03983}~[cs.LG]


\bibitem[\protect\citeauthoryear{Lucas, Sun, Zemel, and Grosse}{Lucas
  et~al\mbox{.}}{2018}]%
        {lucas2018aggregated}
\bibfield{author}{\bibinfo{person}{James Lucas}, \bibinfo{person}{Shengyang
  Sun}, \bibinfo{person}{Richard Zemel}, {and} \bibinfo{person}{Roger Grosse}.}
  \bibinfo{year}{2018}\natexlab{}.
\newblock \showarticletitle{Aggregated momentum: Stability through passive
  damping}.
\newblock \bibinfo{journal}{\emph{arXiv preprint arXiv:1804.00325}}
  (\bibinfo{year}{2018}).
\newblock


\bibitem[\protect\citeauthoryear{Ma and Yarats}{Ma and Yarats}{2018}]%
        {ma2018quasi}
\bibfield{author}{\bibinfo{person}{Jerry Ma} {and} \bibinfo{person}{Denis
  Yarats}.} \bibinfo{year}{2018}\natexlab{}.
\newblock \showarticletitle{Quasi-hyperbolic momentum and {Adam} for deep
  learning}.
\newblock \bibinfo{journal}{\emph{arXiv preprint arXiv:1810.06801}}
  (\bibinfo{year}{2018}).
\newblock


\bibitem[\protect\citeauthoryear{Mikolov, Sutskever, Chen, Corrado, and
  Dean}{Mikolov et~al\mbox{.}}{2013}]%
        {mikolov2013distributed}
\bibfield{author}{\bibinfo{person}{Tomas Mikolov}, \bibinfo{person}{Ilya
  Sutskever}, \bibinfo{person}{Kai Chen}, \bibinfo{person}{Greg~S Corrado},
  {and} \bibinfo{person}{Jeff Dean}.} \bibinfo{year}{2013}\natexlab{}.
\newblock \showarticletitle{Distributed representations of words and phrases
  and their compositionality}. In \bibinfo{booktitle}{\emph{Advances in neural
  information processing systems}}. \bibinfo{pages}{3111--3119}.
\newblock


\bibitem[\protect\citeauthoryear{Mirza and Osindero}{Mirza and
  Osindero}{2014}]%
        {mirza2014conditional}
\bibfield{author}{\bibinfo{person}{Mehdi Mirza} {and} \bibinfo{person}{Simon
  Osindero}.} \bibinfo{year}{2014}\natexlab{}.
\newblock \showarticletitle{Conditional generative adversarial nets}.
\newblock \bibinfo{journal}{\emph{arXiv preprint arXiv:1411.1784}}
  (\bibinfo{year}{2014}).
\newblock


\bibitem[\protect\citeauthoryear{Mitliagkas, Zhang, Hadjis, and
  R{\'e}}{Mitliagkas et~al\mbox{.}}{2016}]%
        {mitliagkas2016asynchrony}
\bibfield{author}{\bibinfo{person}{Ioannis Mitliagkas}, \bibinfo{person}{Ce
  Zhang}, \bibinfo{person}{Stefan Hadjis}, {and} \bibinfo{person}{Christopher
  R{\'e}}.} \bibinfo{year}{2016}\natexlab{}.
\newblock \showarticletitle{Asynchrony begets momentum, with an application to
  deep learning}. In \bibinfo{booktitle}{\emph{2016 54th Annual Allerton
  Conference on Communication, Control, and Computing (Allerton)}}. IEEE,
  \bibinfo{pages}{997--1004}.
\newblock


\bibitem[\protect\citeauthoryear{Nesterov}{Nesterov}{1983}]%
        {nesterov1983nag}
\bibfield{author}{\bibinfo{person}{Yurii Nesterov}.}
  \bibinfo{year}{1983}\natexlab{}.
\newblock \showarticletitle{A method for solving the convex programming problem
  with convergence rate of (1/kˆ2)}.
\newblock \bibinfo{journal}{\emph{Soviet Mathematics Doklady}}
  \bibinfo{volume}{27}, \bibinfo{number}{2} (\bibinfo{year}{1983}),
  \bibinfo{pages}{372--376}.
\newblock


\bibitem[\protect\citeauthoryear{O’donoghue and Candes}{O’donoghue and
  Candes}{2015}]%
        {o2015adaptive}
\bibfield{author}{\bibinfo{person}{Brendan O’donoghue} {and}
  \bibinfo{person}{Emmanuel Candes}.} \bibinfo{year}{2015}\natexlab{}.
\newblock \showarticletitle{Adaptive restart for accelerated gradient schemes}.
\newblock \bibinfo{journal}{\emph{Foundations of computational mathematics}}
  \bibinfo{volume}{15}, \bibinfo{number}{3} (\bibinfo{year}{2015}),
  \bibinfo{pages}{715--732}.
\newblock


\bibitem[\protect\citeauthoryear{Paszke, Gross, Chintala, Chanan, Yang, DeVito,
  Lin, Desmaison, Antiga, and Lerer}{Paszke et~al\mbox{.}}{2017}]%
        {paszke2017automatic}
\bibfield{author}{\bibinfo{person}{Adam Paszke}, \bibinfo{person}{Sam Gross},
  \bibinfo{person}{Soumith Chintala}, \bibinfo{person}{Gregory Chanan},
  \bibinfo{person}{Edward Yang}, \bibinfo{person}{Zachary DeVito},
  \bibinfo{person}{Zeming Lin}, \bibinfo{person}{Alban Desmaison},
  \bibinfo{person}{Luca Antiga}, {and} \bibinfo{person}{Adam Lerer}.}
  \bibinfo{year}{2017}\natexlab{}.
\newblock \showarticletitle{Automatic differentiation in PyTorch}.
\newblock  (\bibinfo{year}{2017}).
\newblock


\bibitem[\protect\citeauthoryear{Radford, Metz, and Chintala}{Radford
  et~al\mbox{.}}{2015}]%
        {radford2015unsupervised}
\bibfield{author}{\bibinfo{person}{Alec Radford}, \bibinfo{person}{Luke Metz},
  {and} \bibinfo{person}{Soumith Chintala}.} \bibinfo{year}{2015}\natexlab{}.
\newblock \showarticletitle{Unsupervised representation learning with deep
  convolutional generative adversarial networks}.
\newblock \bibinfo{journal}{\emph{arXiv preprint arXiv:1511.06434}}
  (\bibinfo{year}{2015}).
\newblock


\bibitem[\protect\citeauthoryear{Rajpurkar, Zhang, Lopyrev, and
  Liang}{Rajpurkar et~al\mbox{.}}{2016}]%
        {rajpurkar2016squad}
\bibfield{author}{\bibinfo{person}{Pranav Rajpurkar}, \bibinfo{person}{Jian
  Zhang}, \bibinfo{person}{Konstantin Lopyrev}, {and} \bibinfo{person}{Percy
  Liang}.} \bibinfo{year}{2016}\natexlab{}.
\newblock \showarticletitle{SQuAD: 100,000+ Questions for Machine Comprehension
  of Text}. In \bibinfo{booktitle}{\emph{Proceedings of EMNLP}} (Austin,
  Texas). \bibinfo{publisher}{Association for Computational Linguistics},
  \bibinfo{pages}{2383--2392}.
\newblock


\bibitem[\protect\citeauthoryear{Reddi, Kale, and Kumar}{Reddi
  et~al\mbox{.}}{2019}]%
        {reddi2019convergence}
\bibfield{author}{\bibinfo{person}{Sashank~J Reddi}, \bibinfo{person}{Satyen
  Kale}, {and} \bibinfo{person}{Sanjiv Kumar}.}
  \bibinfo{year}{2019}\natexlab{}.
\newblock \showarticletitle{On the convergence of {Adam} and beyond}.
\newblock \bibinfo{journal}{\emph{arXiv preprint arXiv:1904.09237}}
  (\bibinfo{year}{2019}).
\newblock


\bibitem[\protect\citeauthoryear{Ren, He, Girshick, and Sun}{Ren
  et~al\mbox{.}}{2015}]%
        {ren2015faster}
\bibfield{author}{\bibinfo{person}{Shaoqing Ren}, \bibinfo{person}{Kaiming He},
  \bibinfo{person}{Ross Girshick}, {and} \bibinfo{person}{Jian Sun}.}
  \bibinfo{year}{2015}\natexlab{}.
\newblock \showarticletitle{Faster {R-CNN}: Towards real-time object detection
  with region proposal networks}. In \bibinfo{booktitle}{\emph{Advances in
  neural information processing systems}}. \bibinfo{pages}{91--99}.
\newblock


\bibitem[\protect\citeauthoryear{Ruder}{Ruder}{2016}]%
        {ruder2016overview}
\bibfield{author}{\bibinfo{person}{Sebastian Ruder}.}
  \bibinfo{year}{2016}\natexlab{}.
\newblock \showarticletitle{An overview of gradient descent optimization
  algorithms}.
\newblock \bibinfo{journal}{\emph{arXiv preprint arXiv:1609.04747}}
  (\bibinfo{year}{2016}).
\newblock


\bibitem[\protect\citeauthoryear{Sabour, Fross, and Hinton}{Sabour
  et~al\mbox{.}}{2017}]%
        {sabour2017caps}
\bibfield{author}{\bibinfo{person}{Sara Sabour}, \bibinfo{person}{Nicholas
  Fross}, {and} \bibinfo{person}{Geoffrey Hinton}.}
  \bibinfo{year}{2017}\natexlab{}.
\newblock \showarticletitle{Dynamic routing between capsules}. In
  \bibinfo{booktitle}{\emph{Advances in neural information processing
  systems}}.
\newblock


\bibitem[\protect\citeauthoryear{Sak, Senior, and Beaufays}{Sak
  et~al\mbox{.}}{2014}]%
        {sak2014long}
\bibfield{author}{\bibinfo{person}{Ha{\c{s}}im Sak}, \bibinfo{person}{Andrew
  Senior}, {and} \bibinfo{person}{Fran{\c{c}}oise Beaufays}.}
  \bibinfo{year}{2014}\natexlab{}.
\newblock \showarticletitle{Long short-term memory recurrent neural network
  architectures for large scale acoustic modeling}. In
  \bibinfo{booktitle}{\emph{Fifteenth annual conference of the international
  speech communication association}}.
\newblock


\bibitem[\protect\citeauthoryear{Sanh, Debut, Chaumond, and Wolf}{Sanh
  et~al\mbox{.}}{2020}]%
        {sanh2020distilbert}
\bibfield{author}{\bibinfo{person}{Victor Sanh}, \bibinfo{person}{Lysandre
  Debut}, \bibinfo{person}{Julien Chaumond}, {and} \bibinfo{person}{Thomas
  Wolf}.} \bibinfo{year}{2020}\natexlab{}.
\newblock \bibinfo{title}{DistilBERT, a distilled version of BERT: smaller,
  faster, cheaper and lighter}.
\newblock
\newblock
\showeprint[arxiv]{1910.01108}~[cs.CL]


\bibitem[\protect\citeauthoryear{Sercu, Puhrsch, Kingsbury, and LeCun}{Sercu
  et~al\mbox{.}}{2016}]%
        {sercu2016very}
\bibfield{author}{\bibinfo{person}{Tom Sercu}, \bibinfo{person}{Christian
  Puhrsch}, \bibinfo{person}{Brian Kingsbury}, {and} \bibinfo{person}{Yann
  LeCun}.} \bibinfo{year}{2016}\natexlab{}.
\newblock \showarticletitle{Very deep multilingual convolutional neural
  networks for {LVCSR}}. In \bibinfo{booktitle}{\emph{2016 IEEE International
  Conference on Acoustics, Speech and Signal Processing (ICASSP)}}. IEEE,
  \bibinfo{pages}{4955--4959}.
\newblock


\bibitem[\protect\citeauthoryear{Shah, Kyrillidis, and Sanghavi}{Shah
  et~al\mbox{.}}{2018}]%
        {shah2018minimum}
\bibfield{author}{\bibinfo{person}{Vatsal Shah}, \bibinfo{person}{Anastasios
  Kyrillidis}, {and} \bibinfo{person}{Sujay Sanghavi}.}
  \bibinfo{year}{2018}\natexlab{}.
\newblock \showarticletitle{Minimum norm solutions do not always generalize
  well for over-parameterized problems}.
\newblock \bibinfo{journal}{\emph{arXiv preprint arXiv:1811.07055}}
  (\bibinfo{year}{2018}).
\newblock


\bibitem[\protect\citeauthoryear{Shankar, Roelofs, Mania, Fang, Recht, and
  Schmidt}{Shankar et~al\mbox{.}}{2020}]%
        {pmlr-v119-shankar20c}
\bibfield{author}{\bibinfo{person}{Vaishaal Shankar}, \bibinfo{person}{Rebecca
  Roelofs}, \bibinfo{person}{Horia Mania}, \bibinfo{person}{Alex Fang},
  \bibinfo{person}{Benjamin Recht}, {and} \bibinfo{person}{Ludwig Schmidt}.}
  \bibinfo{year}{2020}\natexlab{}.
\newblock \showarticletitle{Evaluating Machine Accuracy on {I}mage{N}et}. In
  \bibinfo{booktitle}{\emph{Proceedings of the 37th International Conference on
  Machine Learning}} \emph{(\bibinfo{series}{Proceedings of Machine Learning
  Research}, Vol.~\bibinfo{volume}{119})},
  \bibfield{editor}{\bibinfo{person}{Hal~Daumé III} {and}
  \bibinfo{person}{Aarti Singh}} (Eds.). \bibinfo{publisher}{PMLR},
  \bibinfo{pages}{8634--8644}.
\newblock
\urldef\tempurl%
\url{http://proceedings.mlr.press/v119/shankar20c.html}
\showURL{%
\tempurl}


\bibitem[\protect\citeauthoryear{Sharir, Peleg, and Shoham}{Sharir
  et~al\mbox{.}}{2020}]%
        {sharir2020the}
\bibfield{author}{\bibinfo{person}{Or Sharir}, \bibinfo{person}{Barak Peleg},
  {and} \bibinfo{person}{Yoav Shoham}.} \bibinfo{year}{2020}\natexlab{}.
\newblock \showarticletitle{The Cost of Training NLP Models: A Concise
  Overview}.
\newblock \bibinfo{journal}{\emph{arXiv preprint arXiv:2004.08900}}
  (\bibinfo{year}{2020}).
\newblock


\bibitem[\protect\citeauthoryear{Simonyan and Zisserman}{Simonyan and
  Zisserman}{2014}]%
        {simonyan2014very}
\bibfield{author}{\bibinfo{person}{Karen Simonyan} {and}
  \bibinfo{person}{Andrew Zisserman}.} \bibinfo{year}{2014}\natexlab{}.
\newblock \showarticletitle{Very deep convolutional networks for large-scale
  image recognition}.
\newblock \bibinfo{journal}{\emph{arXiv preprint arXiv:1409.1556}}
  (\bibinfo{year}{2014}).
\newblock


\bibitem[\protect\citeauthoryear{Sivaprasad, Mai, Vogels, Jaggi, and
  Fleuret}{Sivaprasad et~al\mbox{.}}{2019}]%
        {tune_opt}
\bibfield{author}{\bibinfo{person}{Prabhu Sivaprasad}, \bibinfo{person}{Florian
  Mai}, \bibinfo{person}{Thijs Vogels}, \bibinfo{person}{Martin Jaggi}, {and}
  \bibinfo{person}{François Fleuret}.} \bibinfo{year}{2019}\natexlab{}.
\newblock \showarticletitle{On the Tunability of Optimizers in Deep Learning}.
\newblock  (\bibinfo{date}{10} \bibinfo{year}{2019}).
\newblock


\bibitem[\protect\citeauthoryear{Smith}{Smith}{2018}]%
        {smith20181cycle}
\bibfield{author}{\bibinfo{person}{Leslie Smith}.}
  \bibinfo{year}{2018}\natexlab{}.
\newblock \showarticletitle{A disciplined approach to neural network
  hyper-parameters: Part 1 -- learning rate, batch size, momentum, and weight
  decay}.
\newblock \bibinfo{journal}{\emph{arXiv preprint arXiv:1803.09820}}
  (\bibinfo{year}{2018}).
\newblock


\bibitem[\protect\citeauthoryear{Smith, Kindermans, Ying, and Le}{Smith
  et~al\mbox{.}}{2017}]%
        {smith2017increasebatch}
\bibfield{author}{\bibinfo{person}{Samuel Smith}, \bibinfo{person}{Pieter-Jan
  Kindermans}, \bibinfo{person}{Chris Ying}, {and} \bibinfo{person}{Quoc Le}.}
  \bibinfo{year}{2017}\natexlab{}.
\newblock \showarticletitle{Don't Decay the Learning Rate, Increase the Batch
  Size}.
\newblock \bibinfo{journal}{\emph{arXiv preprint arXiv:1711.00489}}
  (\bibinfo{year}{2017}).
\newblock


\bibitem[\protect\citeauthoryear{Socher, Perelygin, Wu, Chuang, Manning, Ng,
  and Potts}{Socher et~al\mbox{.}}{2013}]%
        {socher2013recursive}
\bibfield{author}{\bibinfo{person}{Richard Socher}, \bibinfo{person}{Alex
  Perelygin}, \bibinfo{person}{Jean Wu}, \bibinfo{person}{Jason Chuang},
  \bibinfo{person}{Christopher~D Manning}, \bibinfo{person}{Andrew Ng}, {and}
  \bibinfo{person}{Christopher Potts}.} \bibinfo{year}{2013}\natexlab{}.
\newblock \showarticletitle{Recursive deep models for semantic compositionality
  over a sentiment treebank}. In \bibinfo{booktitle}{\emph{Proceedings of
  EMNLP}}. \bibinfo{pages}{1631--1642}.
\newblock


\bibitem[\protect\citeauthoryear{Song and Ermon}{Song and Ermon}{2019}]%
        {song2019generative}
\bibfield{author}{\bibinfo{person}{Yang Song} {and} \bibinfo{person}{Stefano
  Ermon}.} \bibinfo{year}{2019}\natexlab{}.
\newblock \showarticletitle{Generative Modeling by Estimating Gradients of the
  Data Distribution}.
\newblock \bibinfo{journal}{\emph{arXiv preprint arXiv:1907.05600}}
  (\bibinfo{year}{2019}).
\newblock


\bibitem[\protect\citeauthoryear{Srinivasan, Sankar, and
  Balasubramanian}{Srinivasan et~al\mbox{.}}{2018}]%
        {srinivasan2018adine}
\bibfield{author}{\bibinfo{person}{Vishwak Srinivasan},
  \bibinfo{person}{Adepu~Ravi Sankar}, {and} \bibinfo{person}{Vineeth~N
  Balasubramanian}.} \bibinfo{year}{2018}\natexlab{}.
\newblock \showarticletitle{ADINE: an adaptive momentum method for stochastic
  gradient descent}. In \bibinfo{booktitle}{\emph{Proceedings of the ACM India
  Joint International Conference on Data Science and Management of Data}}. ACM,
  \bibinfo{pages}{249--256}.
\newblock


\bibitem[\protect\citeauthoryear{Sutskever, Martens, Dahl, and
  Hinton}{Sutskever et~al\mbox{.}}{2013}]%
        {sutskever2013importance}
\bibfield{author}{\bibinfo{person}{Ilya Sutskever}, \bibinfo{person}{James
  Martens}, \bibinfo{person}{George Dahl}, {and} \bibinfo{person}{Geoffrey
  Hinton}.} \bibinfo{year}{2013}\natexlab{}.
\newblock \showarticletitle{On the importance of initialization and momentum in
  deep learning}. In \bibinfo{booktitle}{\emph{International conference on
  machine learning}}. \bibinfo{pages}{1139--1147}.
\newblock


\bibitem[\protect\citeauthoryear{Wang, Singh, Michael, Hill, Levy, and
  Bowman}{Wang et~al\mbox{.}}{2019}]%
        {wang2019glue}
\bibfield{author}{\bibinfo{person}{Alex Wang}, \bibinfo{person}{Amanpreet
  Singh}, \bibinfo{person}{Julian Michael}, \bibinfo{person}{Felix Hill},
  \bibinfo{person}{Omer Levy}, {and} \bibinfo{person}{Samuel~R. Bowman}.}
  \bibinfo{year}{2019}\natexlab{}.
\newblock \showarticletitle{{GLUE}: A Multi-Task Benchmark and Analysis
  Platform for Natural Language Understanding}.
\newblock
\newblock
\shownote{In the Proceedings of ICLR.}


\bibitem[\protect\citeauthoryear{Wang, Jiang, Qian, Yang, Li, Zhang, Wang, and
  Tang}{Wang et~al\mbox{.}}{2017}]%
        {wang2017residual}
\bibfield{author}{\bibinfo{person}{Fei Wang}, \bibinfo{person}{Mengqing Jiang},
  \bibinfo{person}{Chen Qian}, \bibinfo{person}{Shuo Yang},
  \bibinfo{person}{Cheng Li}, \bibinfo{person}{Honggang Zhang},
  \bibinfo{person}{Xiaogang Wang}, {and} \bibinfo{person}{Xiaoou Tang}.}
  \bibinfo{year}{2017}\natexlab{}.
\newblock \showarticletitle{Residual attention network for image
  classification}.
\newblock \bibinfo{journal}{\emph{CVPR}} (\bibinfo{year}{2017}).
\newblock


\bibitem[\protect\citeauthoryear{Warstadt, Singh, and Bowman}{Warstadt
  et~al\mbox{.}}{2018}]%
        {warstadt2018neural}
\bibfield{author}{\bibinfo{person}{Alex Warstadt}, \bibinfo{person}{Amanpreet
  Singh}, {and} \bibinfo{person}{Samuel~R. Bowman}.}
  \bibinfo{year}{2018}\natexlab{}.
\newblock \showarticletitle{Neural Network Acceptability Judgments}.
\newblock \bibinfo{journal}{\emph{arXiv preprint 1805.12471}}
  (\bibinfo{year}{2018}).
\newblock


\bibitem[\protect\citeauthoryear{Wibisono and Wilson}{Wibisono and
  Wilson}{2015}]%
        {wibisono2015accelerated}
\bibfield{author}{\bibinfo{person}{Andre Wibisono} {and}
  \bibinfo{person}{Ashia~C Wilson}.} \bibinfo{year}{2015}\natexlab{}.
\newblock \showarticletitle{On accelerated methods in optimization}.
\newblock \bibinfo{journal}{\emph{arXiv preprint arXiv:1509.03616}}
  (\bibinfo{year}{2015}).
\newblock


\bibitem[\protect\citeauthoryear{Wibisono, Wilson, and Jordan}{Wibisono
  et~al\mbox{.}}{2016}]%
        {wibisono2016variational}
\bibfield{author}{\bibinfo{person}{Andre Wibisono}, \bibinfo{person}{Ashia~C
  Wilson}, {and} \bibinfo{person}{Michael~I Jordan}.}
  \bibinfo{year}{2016}\natexlab{}.
\newblock \showarticletitle{A variational perspective on accelerated methods in
  optimization}.
\newblock \bibinfo{journal}{\emph{proceedings of the National Academy of
  Sciences}} \bibinfo{volume}{113}, \bibinfo{number}{47}
  (\bibinfo{year}{2016}), \bibinfo{pages}{E7351--E7358}.
\newblock


\bibitem[\protect\citeauthoryear{Williams, Nangia, and Bowman}{Williams
  et~al\mbox{.}}{2018}]%
        {williams2018broad}
\bibfield{author}{\bibinfo{person}{Adina Williams}, \bibinfo{person}{Nikita
  Nangia}, {and} \bibinfo{person}{Samuel~R. Bowman}.}
  \bibinfo{year}{2018}\natexlab{}.
\newblock \showarticletitle{A Broad-Coverage Challenge Corpus for Sentence
  Understanding through Inference}. In \bibinfo{booktitle}{\emph{Proceedings of
  NAACL-HLT}}.
\newblock


\bibitem[\protect\citeauthoryear{Wilson, Recht, and Jordan}{Wilson
  et~al\mbox{.}}{2016}]%
        {wilson2016lyapunov}
\bibfield{author}{\bibinfo{person}{Ashia~C Wilson}, \bibinfo{person}{Benjamin
  Recht}, {and} \bibinfo{person}{Michael~I Jordan}.}
  \bibinfo{year}{2016}\natexlab{}.
\newblock \showarticletitle{A {L}yapunov analysis of momentum methods in
  optimization}.
\newblock \bibinfo{journal}{\emph{arXiv preprint arXiv:1611.02635}}
  (\bibinfo{year}{2016}).
\newblock


\bibitem[\protect\citeauthoryear{Wilson, Roelofs, Stern, Srebro, and
  Recht}{Wilson et~al\mbox{.}}{2017}]%
        {wilson2017marginal}
\bibfield{author}{\bibinfo{person}{Ashia~C Wilson}, \bibinfo{person}{Rebecca
  Roelofs}, \bibinfo{person}{Mitchell Stern}, \bibinfo{person}{Nati Srebro},
  {and} \bibinfo{person}{Benjamin Recht}.} \bibinfo{year}{2017}\natexlab{}.
\newblock \showarticletitle{The marginal value of adaptive gradient methods in
  machine learning}. In \bibinfo{booktitle}{\emph{Advances in Neural
  Information Processing Systems}}. \bibinfo{pages}{4148--4158}.
\newblock


\bibitem[\protect\citeauthoryear{Wolf, Debut, Sanh, Chaumond, Delangue, Moi,
  Cistac, Rault, Louf, Funtowicz, Davison, Shleifer, von Platen, Ma, Jernite,
  Plu, Xu, Scao, Gugger, Drame, Lhoest, and Rush}{Wolf et~al\mbox{.}}{2020}]%
        {wolf2020huggingfaces}
\bibfield{author}{\bibinfo{person}{Thomas Wolf}, \bibinfo{person}{Lysandre
  Debut}, \bibinfo{person}{Victor Sanh}, \bibinfo{person}{Julien Chaumond},
  \bibinfo{person}{Clement Delangue}, \bibinfo{person}{Anthony Moi},
  \bibinfo{person}{Pierric Cistac}, \bibinfo{person}{Tim Rault},
  \bibinfo{person}{Rémi Louf}, \bibinfo{person}{Morgan Funtowicz},
  \bibinfo{person}{Joe Davison}, \bibinfo{person}{Sam Shleifer},
  \bibinfo{person}{Patrick von Platen}, \bibinfo{person}{Clara Ma},
  \bibinfo{person}{Yacine Jernite}, \bibinfo{person}{Julien Plu},
  \bibinfo{person}{Canwen Xu}, \bibinfo{person}{Teven~Le Scao},
  \bibinfo{person}{Sylvain Gugger}, \bibinfo{person}{Mariama Drame},
  \bibinfo{person}{Quentin Lhoest}, {and} \bibinfo{person}{Alexander~M. Rush}.}
  \bibinfo{year}{2020}\natexlab{}.
\newblock \bibinfo{title}{HuggingFace's Transformers: State-of-the-art Natural
  Language Processing}.
\newblock
\newblock
\showeprint[arxiv]{1910.03771}~[cs.CL]


\bibitem[\protect\citeauthoryear{Xie, Girshick, Doll{\'a}r, Tu, and He}{Xie
  et~al\mbox{.}}{2017}]%
        {xie2017aggregated}
\bibfield{author}{\bibinfo{person}{Saining Xie}, \bibinfo{person}{Ross
  Girshick}, \bibinfo{person}{Piotr Doll{\'a}r}, \bibinfo{person}{Zhuowen Tu},
  {and} \bibinfo{person}{Kaiming He}.} \bibinfo{year}{2017}\natexlab{}.
\newblock \showarticletitle{Aggregated residual transformations for deep neural
  networks}. In \bibinfo{booktitle}{\emph{Proceedings of the IEEE conference on
  computer vision and pattern recognition}}. \bibinfo{pages}{1492--1500}.
\newblock


\bibitem[\protect\citeauthoryear{Yuan, Ying, and Sayed}{Yuan
  et~al\mbox{.}}{2016}]%
        {yuan2016sgdequivalence}
\bibfield{author}{\bibinfo{person}{Kun Yuan}, \bibinfo{person}{Bicheng Ying},
  {and} \bibinfo{person}{Ali Sayed}.} \bibinfo{year}{2016}\natexlab{}.
\newblock \showarticletitle{On the influence of momentum acceleration on online
  learning}.
\newblock \bibinfo{journal}{\emph{Journal of Machine Learning Research}}
  \bibinfo{volume}{17}, \bibinfo{number}{192} (\bibinfo{year}{2016}),
  \bibinfo{pages}{1--66}.
\newblock


\bibitem[\protect\citeauthoryear{Zagoruyko and Komodakis}{Zagoruyko and
  Komodakis}{2016}]%
        {zagoruyko2016wide}
\bibfield{author}{\bibinfo{person}{Sergey Zagoruyko} {and}
  \bibinfo{person}{Nikos Komodakis}.} \bibinfo{year}{2016}\natexlab{}.
\newblock \showarticletitle{Wide residual networks}.
\newblock \bibinfo{journal}{\emph{arXiv preprint arXiv:1605.07146}}
  (\bibinfo{year}{2016}).
\newblock


\bibitem[\protect\citeauthoryear{Zeiler}{Zeiler}{2012}]%
        {zeiler2012adadelta}
\bibfield{author}{\bibinfo{person}{Matthew~D Zeiler}.}
  \bibinfo{year}{2012}\natexlab{}.
\newblock \showarticletitle{ADADELTA: an adaptive learning rate method}.
\newblock \bibinfo{journal}{\emph{arXiv preprint arXiv:1212.5701}}
  (\bibinfo{year}{2012}).
\newblock


\bibitem[\protect\citeauthoryear{Zhang and Mitliagkas}{Zhang and
  Mitliagkas}{2017}]%
        {zhang2017yellowfin}
\bibfield{author}{\bibinfo{person}{Jian Zhang} {and} \bibinfo{person}{Ioannis
  Mitliagkas}.} \bibinfo{year}{2017}\natexlab{}.
\newblock \showarticletitle{Yellowfin and the art of momentum tuning}.
\newblock \bibinfo{journal}{\emph{arXiv preprint arXiv:1706.03471}}
  (\bibinfo{year}{2017}).
\newblock


\bibitem[\protect\citeauthoryear{Zou, Shen, Jie, Zhang, and Liu}{Zou
  et~al\mbox{.}}{2018}]%
        {zou2018a}
\bibfield{author}{\bibinfo{person}{Fangyu Zou}, \bibinfo{person}{Li Shen},
  \bibinfo{person}{Zequn Jie}, \bibinfo{person}{Weizhong Zhang}, {and}
  \bibinfo{person}{Wei Liu}.} \bibinfo{year}{2018}\natexlab{}.
\newblock \showarticletitle{A sufficient condition for convergences of Adam and
  RMSProp}.
\newblock \bibinfo{journal}{\emph{arxiv preprint arXiv:1811.09358}}
  (\bibinfo{year}{2018}).
\newblock


\end{thebibliography}

\begin{appendix}

\clearpage

\section{Experiments} 
\label{detailedExperimentSettings}

We describe the nine test problems in this paper.

\begin{itemize}[leftmargin=*]
  \item \textbf{CIFAR10 - ResNet20.} CIFAR10 contains 60,000 32x32x3 images with a 50,000 training set, 10,000 test set split. There are 10 classes. ResNet20 \cite{he2016deep} is an 20 layers deep CNN with skip connections for image classification. Trained with a batch size of 128.
  \item \textbf{TINY IMAGENET - ResNet56.} Tiny ImageNet contains 110,000 64x64x3 images with a 100,000 training set, 10,000 test set split. There are 200 classes. ResNet56 \cite{he2016deep} is a 56 layer deep CNN with skip connections for image classification. Trained with a batch size of 128.
  \item \textbf{CIFAR100 - VGG16.} CIFAR100 is a fine-grained version of CIFAR-10 and contains 60,000 32x32x3 images with a 50,000 training set, 10,000 test set split. There are 100 classes.
  VGG16 \cite{simonyan2014very} is a 16 layers deep CNN with extensive use of 3x3 convolutional filters. Trained with a batch size of 128.
  \item \textbf{STL10 - Wide ResNet 16-8.} STL10 contains 1300 96x96x3 images with a 500 training set, 800 test set split. There are 10 classes. Wide ResNet 16-8 \cite{zagoruyko2016wide} is a 16 layers deep ResNet which is 8 times wider. Trained with a batch size of 64.
  \item \textbf{PTB - LSTM.} PTB is an English text corpus containing 929,000 training words, 73,000 validation words, and 82,000 test words. There are 10,000 words in the vocabulary. The model is stacked LSTMs \cite{hochreiter1997long} with 2 layers, 650 units per layer, and dropout of 0.5. Trained with a batch size of 20. We use the official TensorFlow v1 implementation for PTB - LSTM.
  \item \textbf{FMNIST - CAPS.} FMNIST contains 60,000 32x32x1 grayscale images with a 50,000 training set, 10,000 test set split. There are 10 classes of 10 clothing items.
  Capsule Networks \cite{sabour2017caps} represent Neural Networks as a set of capsules, where each capsule encodes a specific entity or meaning. The activations of capsules depend on comparing incoming pose predictions, as opposed to standard neural networks. The Capsule Network uses 3 iterations in the routing algorithm. Trained with a batch size of 128.
  \item \textbf{MNIST - VAE.} MNIST contains 60,000 32x32x1 grayscale images with a 50,000 training set, 10,000 test set split. There are 10 classes of 10 digits. VAE \cite{kingma2015vae} with three dense encoding layers and three dense decoding layers with a latent space of size 2.
  Trained with a batch size of 100.
  \item \textbf{CIFAR10 - NCSN.} CIFAR10 contains 60,000 32x32x3 images with a 50,000 training set, 10,000 test set split. There are 10 classes. NCSN \cite{song2019generative} is a recent state-of-the-art generative model which achieves the best reported inception score. We compute inception scores based on a total of 50000 samples. Since \textsc{Demon} depends on a predefined number of epochs, we evaluate inception score at the end of training; otherwise, we follow the exact implementation in and defer details to the original paper.
  \item \textbf{GLUE - BERT.} The GLUE benchmark \cite{wang2019glue} consists of 9 different language tasks \cite{warstadt2018neural, socher2013recursive, dolan2005automatically, agirre2007semantic, williams2018broad, rajpurkar2016squad, dagan2006pascal, bar2006second, giampiccolo2007third, bentivogli2009fifth, levesque2011winograd}, grouped together to form a benchmark. BERT \cite{devlin2019bert} is a relatively recently proposed language model which has become the standard for many tasks in NLP. In particular, BERT can be fine-tuned to an array of tasks, and here we evaluate the fine-tuning procedure of BERT to the GLUE benchmark.
\end{itemize}

\end{appendix}
\end{document}